\documentclass{article}

\usepackage{microtype}
\usepackage{graphicx}
\usepackage{subcaption}
\usepackage{booktabs}
\usepackage{hyperref}

\usepackage[preprint]{icml2026}

\usepackage{amsmath,amssymb,mathtools,amsthm}
\usepackage[capitalize,noabbrev]{cleveref}
\usepackage[textsize=tiny]{todonotes}
\usepackage{colortbl}
\usepackage{multirow}
\usepackage{hhline}
\definecolor{tabframe}{RGB}{90,110,140}
\definecolor{tabhead}{RGB}{214,224,238}
\definecolor{tablabel}{RGB}{238,242,248}

\newcommand{\profile}{\varphi}
\newcommand{\weights}{w}
\newcommand{\interest}{I}
\newcommand{\mathlib}{Mathlib}

\newcommand{\Lmain}{\textsc{Marked}}
\newcommand{\Llists}{\textsc{Listed}}
\newcommand{\Lnamed}{\textsc{Named}}
\newcommand{\Lintro}{\textsc{Stated}}

\icmltitlerunning{What makes a theorem interesting?}

\begin{document}

\twocolumn[
  \icmltitle{What makes a theorem interesting?\\
    Operationalizing interestingness for open-ended exploration}

  \begin{icmlauthorlist}
    \icmlauthor{Silv\`ere Gangloff}{aff}
  \end{icmlauthorlist}
  \icmlaffiliation{aff}{University of Ostrava, Ostrava, Czech Republic}
  \icmlcorrespondingauthor{Silv\`ere Gangloff}{sfgangloff@gmail.com}
  \icmlkeywords{AI for mathematics, interestingness, formal mathematics, Lean, Mathlib}
  \vskip 0.3in
]

\printAffiliationsAndNotice{}

\begin{abstract}
An agent that explores, whether it generates conjectures or research ideas, must decide which of them are worth keeping. This judgment of interestingness is usually delegated to a hand-picked proxy or to a language model, neither of which is known to correspond to what humans actually value. I study it for mathematical theorems, by observing which results mathematicians single out in formal libraries and in papers. I model each way of singling out results as an evaluator that weights features of a result's statement, its proof and dependencies, and split its weights into a shared and a subjective part. In formalized mathematics, ex-ante features predict these selections as well as a large language-model judge, and the two complement each other. Interestingness is also largely shared. One set of weights, carried mostly by structural features such as the length of the statement and of the proof and the subfields the proof draws on, accounts for most of the signal and transfers in part across formal libraries. What remains subjective depends on practices and individual authors more than on subfields. The judge's advantage lies in its knowledge of the canon, and neither the judge nor the features rely on memory of the papers. The approach, which infers what a community finds interesting from what it singles out, can be applied to other sciences.
\end{abstract}

\section{Introduction}
\label{sec:intro}

In a few years, language models have evolved from solving high-school competition problems \citep{trinh2024alphageometry} to contributing
to research mathematics, on open problems as well as on the formalization of known proofs. In
2026 alone, an internal OpenAI model disproved
Erd\H{o}s's unit distance conjecture, a counterexample since checked and digested by
mathematicians \citep{alon2026unitdistance}, and
Claude formalized Fermat's Last Theorem in Lean, largely autonomously, in eleven days
\citep{anthropic2026flt}.

All these contributions were produced within a relatively fixed conceptual framework, in pursuit of a fixed objective stated in it. Mathematicians also explore without a
fixed objective. They choose which questions to formulate, which
definitions to write down and which results deserve to be stated, and these choices reshape the framework
itself. Models do not yet explore in this way, as they have not mastered the abductive leap that produces new meaningful premises \citep{zahavy2026jump} (\cref{app:framework}).

For an agent to explore mathematics open-endedly and produce something of value, it must know
what humans find interesting. Mathematics is not meant to only record everything that is true, which a machine could enumerate \citep{newell1957logic}, but to produce a minimal conceptual framework from which existing mathematics can be rebuilt, and a result is interesting, from the point of view of pure mathematics, if it earns its place in that framework (\cref{app:framework}). Note that pure mathematics is only one point of view, and scientists who need mathematics value it for what it lets them model, compute or predict \citep{hardy1940apology,wigner1960effectiveness}.

These considerations lead to the central question of this paper. How can a model learn what mathematicians find
interesting? An exploring agent faces it at every step. Which conjectures should it try to prove
first? When it formulates a new one, is it interesting enough to share? The question is as old as automated discovery (\cref{sec:related}), and agents for research and for mathematics now generate ideas, conjectures and lemmas faster than anyone can read them. As
generation becomes cheap, selection becomes the bottleneck.

Most current systems rely on a single proxy, such as the ratio of proof length to statement length, compression or the verdict of a language model (\cref{sec:related}). The few that learn a measure of interestingness learn it from their own exploration \citep{schmidhuber2010creativity,tsoukalas2025learning}, independently of what humans select. Each proxy captures part of the notion, but none has been measured against what
mathematicians single out, and each has obvious counterexamples when used alone (\cref{app:framework}). Nor is it
clear whether a language-model judge measures the result itself or its memory of results that
became known. It is not even clear that interestingness is one quantity, nor how much of it is shared and how much belongs to who judges.

\paragraph{Contributions.} I study the question in mathematics, where it can be measured, since a
result, its proof and what it depends on are explicit, and fully formal in proof assistants. I focus on theorems, study whether what mathematicians find interesting can be learned before anyone uses a result, and measure how much of it is shared among evaluators (\cref{app:framework}). I make four contributions.
\begin{enumerate}
  \item \textbf{A conceptual framework} (\cref{sec:framework}). I define evaluators, profiles and the split of interestingness into a shared and a subjective part, which turns vague questions about interestingness into testable ones.
  \item \textbf{A measurement of revealed preferences} (\cref{sec:framework,sec:data}). I observe what mathematicians single out in formal libraries and in papers, which I call \emph{discourse centrality}, and argue that it is a reasonable proxy for interestingness.
  \item \textbf{Unexpected answers.} Two of them stand out: what mathematicians select is known ex ante, as well as a language-model judge, and the two complement each other (\cref{sec:prediction}); interestingness is largely shared, and, contrary to intuition, subfields have no weights of their own, while practices and individual authors do (\cref{sec:core,sec:libraries}).
  \item \textbf{Data for evaluating} any agent's notion of interestingness. I release the extraction,
    features, labels and protocols for \mathlib{} \citep{mathlib2020}, Metamath \citep{metamath},
    HOL Light \citep{harrison2009hollight} and 20k arXiv papers.
\end{enumerate}
\section{Conceptual framework}
\label{sec:framework}

\paragraph{Profiles of results.} Let $x$ be a mathematical result, a statement together with a proof in a
body of mathematics $\mathcal{L}$ (a library or a literature), and let $\mathcal{X}_\mathcal{L}$
be the set of results of $\mathcal{L}$. A \emph{feature} is a computable function
$\profile_k(\,\cdot\,; \mathcal{L}) : \mathcal{X}_\mathcal{L} \to \mathbb{R}$ that measures one
aspect of a result. The features I use are listed in \cref{sec:data}. The \emph{profile} of $x$ is its feature vector
$\profile(x) = (\profile_1(x), \ldots, \profile_K(x)) \in \mathbb{R}^K$.
Existing proxies, such as the proof/statement ratio of \citet{patel2026interesting},
compression gain in the sense of \citet{bengio2024aimathematician}, usage counts and the
surprisal of the statement to a language model, are single features.

\paragraph{Evaluators and their value functions.} I call \emph{evaluator} any agent $c$ that selects results according
to a value function $\interest_c : \mathcal{X}_\mathcal{L} \to \mathbb{R}$. Faced with several
results, $c$ singles out those of highest value first. The value function of $c$ is defined as
\begin{equation}
  \interest_c(x) = f\big(\weights_c, \profile(x)\big),
  \label{eq:value}
\end{equation}
where $\weights_c \in \mathbb{R}^K$ contains one weight per feature, saying how much $c$ cares about each aspect, and $f$ combines these weights with the profile into a value. Only the order induced by $\interest_c$ matters (\cref{app:framework}).

\paragraph{Objective and subjective evaluations.} Fix a family $\mathcal{C}$ of evaluators.
The evaluators I consider are all human, and mostly collective. Most of them are \emph{signallers}, groups of
people defined by a practice through which they single out results. Examples are formalizers marking the main results of a file, curators compiling lists, the community naming results, and authors choosing what to state in their introduction (\cref{app:framework}).
I also treat as evaluators the \emph{subfields}, the communities working in one area of mathematics, such as
algebra or probability, since the habits of an area may shape what
its members value. On arXiv,
I can also treat each \emph{author} as an evaluator. I split the
weights of each evaluator into the center of the family and a deviation from it. After normalizing each weight vector to Euclidean norm 1 (\cref{app:framework}), I set
\begin{equation}
  \weights_0 = \frac{1}{|\mathcal{C}|} \sum_{c \in \mathcal{C}} \weights_c, \qquad
  \delta_c = \weights_c - \weights_0, \qquad c \in \mathcal{C}.
  \label{eq:decomposition}
\end{equation}

\paragraph{Shared interestingness and objectivity.} I say that the center is \emph{shared}
by the family when each evaluator loses little by using it in place of its own weights, that is,
when the center ranks results nearly as the evaluator does. A center can be computed for any family,
so whether it is shared is a purely empirical question. I measure how little each evaluator loses (\cref{app:framework} provides an example) by the \emph{share} of the center,
\begin{equation}
  s_c = \frac{\mathrm{AUROC}_c(\weights_0) - \tfrac12}{\mathrm{AUROC}_c(\weights_c) - \tfrac12},
  \label{eq:share}
\end{equation}
the fraction of its above-chance signal kept when its own weights are replaced by the center.
Here $\mathrm{AUROC}_c(\weights)$ is the probability that the score
$\langle \weights, \profile(x) \rangle$ ranks a result singled out by $c$ above a result it did not single out ($\tfrac12$ corresponds to
chance, 1 to a perfect ranking) \citep{hanley1982meaning}. When $s_c$ is close to 1 for every $c \in \mathcal{C}$, I call the
value function $\interest_0 = f(\weights_0, \profile(\cdot))$ the \emph{shared interestingness} of $\mathcal{C}$, and the
deviations $\delta_c$ its \emph{subjective part}, which I also call the \emph{taste} of $c$. 

Shared interestingness is not yet objective. Agreement among evaluators of one culture is evidence of \emph{intersubjectivity} only, since it may reflect what they have learned to value together. I therefore widen the family progressively, to other libraries, informal mathematics and later evaluators, and call shared interestingness \emph{more objective} the further it holds (\cref{app:framework}). Finally, shared interestingness is only
as rich as the features, since what the profile cannot express ends up in the deviations or in the
noise.

\paragraph{Assumption of linearity.} In this paper I assume that each value function is linear in the features. I write $\interest_c(x) = \langle \weights_c, \profile(x) \rangle$, and use as features the percentiles of the raw measurements within each subfield (\cref{app:framework}).

\paragraph{Interestingness and discourse centrality.} The weights $\weights_c$, and hence $\interest_c(x)$, are not observed (\cref{app:framework}). Among what I can measure is
\emph{discourse centrality}, that is, whether $c$ has put the result at the center of its discourse (a paper, a formal library, a textbook or a list of
theorems), by
marking it as a main result, listing it, naming it, stating it in an introduction or citing it.
It differs from \emph{dependency centrality}, how many other results use a result in their proofs, which is a feature (\cref{app:framework}). For each evaluator $c$, I encode discourse centrality as a binary \emph{label} $y_c(x)$, equal
to 1 if $c$ has singled out $x$ and to 0 otherwise. \Cref{sec:data} lists all labels. 

Since $y_c$ is binary and $\interest_c$ real, the label cannot stand in for interestingness value by value. What
links them is an assumption on order, which I call \emph{monotone selection}, as in random-utility models of choice \citep{mcfadden1974conditional}: the more
interesting a result is for $c$, the more likely $c$ is to single it out. Formally, there is an
increasing function $F_c$ such that
\begin{equation}
  \Pr\big[y_c(x) = 1\big] = F_c\big(\interest_c(x)\big).
  \label{eq:monotone}
\end{equation}
The label is thus a coarse and noisy trace of the value function, since $c$ singles out $x$ roughly when
$\interest_c(x)$ is high.  Since $\interest_c$ matters only through the order it induces, this is
all I need. Under \eqref{eq:monotone}, the results $c$ singled out tend to rank above the others,
so the labels reveal the order of $\interest_c$, up to noise, though not its values. This is the
precise sense in which I use discourse centrality as a proxy for interestingness.

\Cref{app:framework} discusses why this assumption is reasonable, where it fails, and why discourse centrality remains a reasonable proxy. The labels overlap little, so I report them separately.

I estimate $\weights_c$ by logistic regression, which takes
$F_c(t) = \sigma(b_c + t)$ in \eqref{eq:monotone}, with $\sigma(t) = 1/(1+e^{-t})$ and an
intercept $b_c$ that absorbs how often $c$ singles out anything. 
In \cref{sec:core} I fit $\weights_0$ and all $\delta_c$ jointly, with a ridge penalty that keeps each $\delta_c$ small unless the data require it (\cref{app:framework}).

\paragraph{Questions on interestingness.} Several intuitive ideas about interestingness can be expressed in this
framework as precise questions, which are the object of the experiments in the rest of the paper:
\begin{itemize}
  \item[(i)] \emph{Is interestingness visible in the result and what it builds on?} (\cref{sec:prediction})
  \item[(ii)] \emph{How much of interestingness is shared?} (\cref{sec:core})
  \item[(iii)] \emph{How objective is shared interestingness?} (\cref{sec:libraries})
  \item[(iv)] \emph{Can a model trained on past selections anticipate later ones?} (\cref{sec:forward})
  \item[(v)] \emph{What lies beyond the current features?} (\cref{sec:limits})
\end{itemize}
\Cref{app:framework} explains where each question comes from.
\section{Data and operationalization}
\label{sec:data}

\paragraph{The \mathlib{} corpus.} \mathlib{} \citep{mathlib2020} is the main library of formalized
mathematics for the Lean proof assistant, consisting of a single, collaboratively written body of definitions and
theorems, each with a machine-checked proof, organized into files (\emph{modules}) and directories
by area. I use the version tagged v4.34.1. I keep only the hand-written theorems, excluding the auxiliary declarations that Lean generates, the theorems of infrastructure directories and instances (\cref{app:extraction}). A result's \emph{subfield} is the
top-level directory it lies in (Algebra, Analysis, Topology, NumberTheory, Probability, \ldots). The other corpora I use serve mainly to widen the family of evaluators and are described in
\cref{sec:libraries}.

\paragraph{Evaluators and their labels.} In \mathlib{} I use three signallers. Each
leaves one label, named after what it records.
\begin{itemize}
  \item \textbf{\Lmain}. The contributors who write \mathlib{} begin each module
    with documentation that lists its main results under a heading ``Main results'' or ``Main
    statements''. The label of a theorem is 1 if the documentation of its module lists it among these main
    results, and 0 otherwise. I call the theorems with label 1 the \emph{positives} of the label;
    there are 3{,}205 of them, spread over 17 subfields.
  \item \textbf{\Llists}. A theorem is positive for this label if it is linked from one of four lists
    maintained alongside \mathlib{}, such as Wiedijk's \emph{100 theorems} (\cref{app:extraction}).
  \item \textbf{\Lnamed}. A theorem is positive for this label if the declaration's docstring calls it by a name (``Fermat's little
    theorem'', ``the Cauchy--Schwarz inequality''; 1{,}684 positives). Restricted to names after
    a person (1{,}098 eponymous results), every result below holds (\cref{app:extraction}).
\end{itemize}
The labels barely overlap (Jaccard index 0.03 to 0.10, \cref{app:extraction}).

I chose these evaluators because they speak for their community, their labels are recorded at scale and independently of my study, and they differ in scope, time, who decides and medium; \cref{app:framework} discusses this choice and its limits.

\paragraph{Features of a result.} I call \emph{ex-ante} the features that can be computed as soon as a result is stated and proved, and \emph{ex-post} those that depend on how later results use it in their proofs (definitions in \cref{app:facets}).
\emph{Statement} features measure the size of the statement, its arguments by kind, its surprisal to two language models and its novelty with respect to earlier declarations. \emph{Proof} features measure the height of the dependency tree below the theorem, the size of its proof, the proof/statement ratio \citep{patel2026interesting}, the entropy of the subfields its dependencies come from, and its novelty with respect to them. \emph{Usage} features, or dependency centrality, measure how much later results use the theorem.
Language models never see the names of theorems, which may reveal how well known they are.

\paragraph{Language-model judges.} In practice, whether a result is interesting is most often decided by a language model
\citep{zhang2024omni,long2026meca}, a practice known as LLM-as-a-judge \citep{zheng2023judging}. I use such a judge as a baseline, as a probe of what the profile misses, and as an evaluator whose value function I can query directly (\cref{app:mathlib}). The judge is not trained on my labels. \emph{Qwen2.5-32B-Instruct} \citep{qwen2024qwen25}, and for robustness \emph{Llama-3.1-70B-Instruct} \citep{grattafiori2024llama3}, rates how significant working mathematicians would consider a result, from its statement or from the whole declaration with its proof (\cref{app:mathlib}).
I score it like profiles, and add its score to the features to test whether the judge and the features are complementary.

\paragraph{Training and evaluation.} For each label I fit the logistic regression of
\cref{sec:framework} on the within-subfield percentiles of the features. I use 5-fold cross-validation grouped by module \citep{roberts2017crossvalidation}, so that
results of one module, which are written together and resemble each other, are never split between
training and evaluation.

I measure how well any score, such as a fitted value function, a single feature or the judge's rating, orders results by the AUROC (\cref{sec:framework}). I compute it within each subfield and average over subfields (\cref{app:mathlib}). An AUROC of 0.78 for \Lmain{} thus means that, within a subfield, a theorem listed
as a main result is ranked above an unlisted one in 78\% of pairs. Confidence intervals are 95\% bootstrap intervals over modules (\cref{app:mathlib}). I write an estimate followed by its interval in brackets, as in $+0.019$ $[+0.009, +0.033]$.
\section{Can interestingness be learned ex ante?}
\label{sec:prediction}

I address question (i) of \cref{sec:framework}: is interestingness visible in a result and what
it builds on? If it is, the profile of a theorem should predict whether each \mathlib{} evaluator
singles it out, using only features available once the theorem is proved. I also compare the profile with single features, with later usage and with a language-model judge.

\begin{table}[t]
\centering
\caption{How well each predictor orders \mathlib{} theorems for each evaluator (within-subfield
AUROC). Brackets:
95\% bootstrap intervals over modules. The judge shown the proof is in \cref{tab:e1proof}.}
\label{tab:e1}
\small
\setlength{\tabcolsep}{2.4pt}
\begin{tabular}{@{}lccc@{}}
\toprule
Predictor & \Lmain & \Llists & \Lnamed \\
\midrule
Proof/statement ratio & 0.685 & 0.666 & 0.693 \\
Dependency height & 0.725 & 0.715 & 0.708 \\
Usage features (ex-post) & 0.581 & 0.630 & 0.634 \\
\textbf{Ex-ante} & \textbf{0.783} & \textbf{0.833} & \textbf{0.830} \\
 & {\scriptsize[0.768,0.799]} & {\scriptsize[0.800,0.857]} & {\scriptsize[0.808,0.849]} \\
Judge 32B, statement & 0.759 & 0.814 & 0.767 \\
Ex-ante + judge 32B & 0.813 & 0.871 & 0.844 \\
\bottomrule
\end{tabular}
\end{table}

\Cref{tab:e1} reports the results, as within-subfield AUROCs (\cref{sec:data}).

\paragraph{Interestingness does not reduce to any single feature.} Ranking theorems by a single feature always does worse than ranking them by the whole profile (\cref{tab:e1}). Interestingness is thus not one quantity that a well-chosen proxy could
capture. The formalizers' main results are mostly readable from the proof, whereas results put on lists or given a name are already largely readable from the statement (\cref{app:mathlib}).

\paragraph{Later usage adds no information.} The ex-post features measure the dependency centrality of a
theorem once the library has grown around it (\cref{sec:data}). I do not include the judges here. Adding
these features to the ex-ante profile changes AUROC by $+0.000$, $+0.003$ and $-0.001$, with every interval
containing 0. For theorems, what evaluators single out is predictable as soon as a result is
proved, before anyone uses it, which is what an exploring agent needs. 

\paragraph{Shown the proof, the judge does as well as the ex-ante features, except on curated lists, where it does better.} The comparison is fair
only if the judge sees what the features see. Shown only the statement, the 32B judge is at a
disadvantage; shown the whole declaration with its proof, on a subset of about 24{,}000 theorems,
it gains $+0.03$ to $+0.05$ on every label. The features then tie with this judge on \Lmain{} and \Lnamed{} and trail it on \Llists{} ($-0.045$ $[-0.091, -0.009]$).

\paragraph{What the judge knows is the canon.} Why does the judge lead on curated lists? Not because it remembers well-known statements, since its lead is the same on every list, the undergraduate curriculum included, and it \emph{trails} the features on named results ($-0.023$ $[-0.042, -0.007]$). What it knows is the \emph{canon}, the results that are taught, surveyed and listed as landmarks, which no profile of the result can contain (\cref{app:mathlib}).

\paragraph{Judges and features are complementary.} The features add
$+0.029$, $+0.014$ and $+0.041$ to the judge shown the proof, and $+0.004$ to $+0.05$ to the 70B
judge. The two judges agree closely with each other (\cref{app:mathlib}). Features and judges capture partly
different things, and an agent would gain from using
both.

The position of a theorem in its file, the presence of documentation, and module documentation seen through surprisal in context explain none of these results (\cref{app:mathlib}).

\paragraph{Answer to question (i).} For theorems in \mathlib{}, what each
evaluator singles out is largely visible in the result and what it builds on, before anyone uses it, and a profile of
it matches a language-model judge except where knowledge of the canon is needed.
\section{How much of interestingness is shared?}
\label{sec:core}

I address here question (ii) of \cref{sec:framework}, how much of interestingness is shared. 

\paragraph{The shared weights carry most of the signal.} I fit one model to the three signallers
of \mathlib{} jointly, with shared weights $\weights_0$ and one deviation $\delta_c$ per signaller,
kept small by a ridge penalty (\cref{app:framework}). Ranking each signaller's results with the
shared weights alone yields an AUROC of 0.758 on \Lmain{}, 0.829 on \Llists{} and 0.822 on
\Lnamed{}. Adding each signaller's deviation raises these to 0.783, 0.833 and 0.830, the same
values as a separate model fitted to each label. The share of the shared weights is therefore
$s_c = 91\%$ for the formalizers (\Lmain{}), $99\%$ for the curated lists (\Llists{}) and $98\%$ for the community (\Lnamed{}). The deviations always help, so the subjective part exists, but it
is small.

\paragraph{Subfields have no weights of their own.} One might expect each subfield of mathematics to value
different aspects (\cref{sec:framework}). To test this, I provide each subfield with its own weights, fitted
only on the results of that subfield, and compare them with the weights pooled over all subfields. If subfields had distinct tastes, their own weights should do better. However, they
do worse. Pooled minus own-subfield AUROC is $+0.019$ $[+0.009, +0.033]$ on \Lmain{}, $+0.026$
$[-0.000, +0.051]$ on \Llists{} and $+0.010$ $[-0.001, +0.023]$ on \Lnamed{}. Subfields at the border of mathematics do not differ more than other groups of subfields either (\cref{app:core}). The same
holds across about 30 arXiv categories (\cref{sec:libraries}). The intuition that subfields value different aspects does not hold for what my features can express.

\paragraph{Signallers deviate from the shared weights.} I train the
model on the label of one signaller and use it to rank results for another (\cref{tab:transfer}).
Weights trained on another label always do worse than weights trained on the label itself, by
$+0.018$ to $+0.077$ AUROC, and every interval excludes 0. They persist when I restrict to documented theorems (\cref{app:core}). The gaps thus reflect real differences between the signallers, which subfields do not have.

\paragraph{Individual authors have weights of their own.} On arXiv, each author is an evaluator through the results they state in their introductions (\cref{sec:libraries}). Deviations estimated on two halves of an author's papers agree four to six times more than two random sets of papers from the author's main category, so individual taste is real and stable, and it is small compared to the shared weights (\cref{app:core}).

\paragraph{The shared weights favour a deep and broad proof, long relative to its statement.}
\Cref{fig:core} (\cref{app:core}) shows the shared weights $\weights_0$ of the joint model above, fitted on
standardized features, with the 95\% interval of each shared weight. For 13 of the 15 features, the
interval lies entirely on one side of 0, so the sign of the shared weight does not depend on which modules
happen to be in the library.  The shared weights raise the score when the proof is long relative to
the statement and rests on a deep chain of results drawn from several subfields, when the result is
new with respect to earlier ones, and when its hypotheses and structural assumptions are written
out. They lower it when many arguments are left for Lean to infer, when the statement surprises a
general language model, and when the result is close to those its proof uses. Only the signs of the shared weights can be interpreted with confidence, and each of them agrees with the effect of its feature taken alone (\cref{app:core}).

\paragraph{Where signallers diverge.} The signallers' own weights differ most on statement size, dependency height, novelty, surprisal and structural assumptions (\cref{fig:core}). For instance, curated lists favour results deep in the dependency tree, and the community names results with rich statements that look familiar to a language model (\cref{app:core}). Part of these differences may come from what each signaller is meant to contain, and only the deviations of individual authors, who all leave the same kind of label, establish that the subjective part is partly a matter of taste (\cref{app:core}).

\paragraph{Answer to question (ii).} Interestingness is largely shared. In \mathlib{}, one set of
weights keeps 91--99\% of each signaller's signal. Giving each subfield its own weights does not improve prediction, while practices
and individual authors have small, stable deviations.
\section{Is interestingness shared across libraries?}
\label{sec:libraries}

I address question (iii) of \cref{sec:framework}, whether the findings of \cref{sec:core} on the
family of evaluators extracted from \mathlib{} extend beyond it. I test two of them: that subfields have
no weights of their own, and that the shared weights predict what other communities single out. \mathlib{} is written and curated by one community, and this may be why
its subfields share one set of weights.  I therefore widen the family.
Informal mathematics on arXiv brings many independent communities, but no formal proofs. Two
other proof assistants, Metamath and HOL Light, bring other communities, other logics and other
conventions, with formal proofs.

\subsection{Informal mathematics on arXiv}
\label{sec:arxiv}

\paragraph{The arXiv corpus and its label.} I extract the theorems, propositions, corollaries and lemmas of 20{,}144 arXiv papers from 2010--2018 (384k results), with their proofs where they can be found (\cref{app:arxiv}). The arXiv categories play the role of the subfields of \mathlib{}. I keep the features that carry over from \mathlib{}, and lose those that need dependencies between papers, such as the entropy of the subfields a proof draws on and usage. The label
\Lintro{} records the choices of the authors. A theorem, proposition or corollary is positive if it
is stated in the introduction of the paper, and it is compared with the other results of the same
paper, after removing restatements. AUROCs are computed within each paper (\cref{app:arxiv}).

\paragraph{Categories have no weights of their own.} As with \mathlib{}'s subfields, weights fitted
separately for each category do no better than pooled weights ($-0.001$ $[-0.002, +0.000]$ across
about 30 categories). The categories at the border of mathematics do not have weights of their own either (\cref{app:arxiv}).

\paragraph{Without formal proofs, the features are weaker than the judge.} The statement features
alone reach an AUROC of only 0.529 $[0.525, 0.534]$ on \Lintro{}, barely above chance. The judge
reaches 0.669 $[0.665, 0.673]$, and combining it with the statement features does not improve its AUROC. As in \mathlib{},
the proof carries the most informative computable features. The proof features reach 0.626, the height of the internal dependency chain alone reaches
0.620, and together they add $+0.017$ $[+0.010, +0.023]$ to the judge. The gap comes from how arXiv differs from \mathlib{}. The dependencies of an informal proof are mostly implicit, and the
features that need them are missing.

\subsection{Metamath and HOL Light}

\paragraph{The two libraries and their labels.} I use the set.mm library of Metamath
\citep{metamath} and the library of HOL Light \citep{harrison2009hollight}. In each library, the labels are the results of Wiedijk's ``100 theorems'' list formalized there, the results whose comment names them, but not after a person, and the results named after a person. In set.mm, I also use the results whose comment cites a textbook
(\cref{app:libraries}). I compare each library's own model with the shared weights $\weights_0$ of \mathlib{}, fitted in \mathlib{} on the five features the three libraries share (\cref{tab:libraries}).

\paragraph{The shared direction transfers to curated results.} The shared weights of \mathlib{}
predict the ``100 theorems'' list of set.mm almost as well as set.mm's own model does (0.769 against
0.778), with the same sign on every feature. They predict the results of HOL Light named after a
person better than HOL Light's own model (0.749 against 0.715). They transfer only partly to HOL Light's ``100 theorems'' list (\cref{app:libraries}). The three libraries agree on two of the five shared features. In all three, a result is more likely to be curated when its proof is long compared with its statement, and when it rests on a long chain of earlier results (\cref{app:libraries}).

\paragraph{The shared weights transfer to results named after a person, but not to other named results.} In set.mm, results named after a person are mostly substantial theorems, such as Sylow's theorems, whereas results with other names are mostly elementary laws, such as the commutative law (\cref{app:libraries}). The shared weights of \mathlib{} predict the former better than set.mm's own model (0.672 against 0.654), but rank the latter below the other results (AUROC 0.416), and no better than at random in HOL Light (0.536). Yet each library's own model predicts them (0.691 and 0.759), so these results follow weights of their own. The signallers differ accordingly. In both libraries, the largest gaps between matched and mismatched labels involve the results with other names (\cref{app:libraries}).

\paragraph{Answer to question (iii).} The absence of subfield-specific weights, across about 30 arXiv categories, and the main direction of the shared weights, for curated results and results named after a person in three proof assistants, extend to other communities. This part of shared interestingness is therefore more objective, in the sense of \cref{sec:framework}, than the agreement within one library could establish. The results that a community names, but not after a person, follow weights of their own.
\section{Can selections be anticipated?}
\label{sec:forward}

I address question (iv) of \cref{sec:framework}, whether a model that learned from information up to some date can anticipate the selections made after it. Many labels were set long after the result, so a model may recognize how well known a result is instead of how interesting it is (\cref{app:forward}). I therefore predict selections made after the
information my models learned from, and report every result for the profile and for the judge.

\paragraph{Profiles predict additions to curated lists.} From \mathlib{}'s git history I
recover, at four cut dates, which theorems existed and what the four curated lists contained
(details in \cref{app:forward}). A model trained on the lists as they stood at a cut ranks the theorems that existed then and were not listed, and a theorem is positive if it was added later. A check on the theorems whose source text did not change after the cut rules out hindsight in the features (\cref{app:forward}). \Cref{tab:x4} reports the results. The
profile ranks the theorems added up to 2.5 years later at a median 87th to 89th percentile (AUROC
0.80 to 0.85). The judge does better. The later additions are mostly part of the canon, which the judge knows (\cref{sec:prediction}), so this test forecasts curation more than the course of mathematical research (\cref{app:forward}).

\paragraph{Profiles anticipate which results later papers cite.} Citing a result is a
selection made by other authors, years after the paper was written. To observe such selections, I use the arXiv papers of \cref{sec:arxiv}. I keep the sentences of later papers that cite one of them together with the number of one of its results, as in ``[12, Theorem 3.2]''. In the 4{,}244 papers that have both cited and uncited results (\cref{app:forward}), within each paper I compare the cited results with the uncited ones. I find the following:
\begin{itemize}
  \item The authors' own choices predict later citations: being stated in the introduction and the kind of result together reach an AUROC of 0.625 $[0.616, 0.632]$. I therefore use these choices as a baseline.
  \item Over all results, although neither sees the section or the kind of result, the profile alone reaches 0.569 $[0.561, 0.577]$, the judge alone 0.591, and the two together 0.601 $[0.593, 0.612]$, and they add
    $+0.025$ $[+0.020, +0.030]$ to the authors' choices.
  \item Citing authors value nearly the same properties of a result as the authors, like the signallers of \mathlib{} (\cref{app:forward}).
\end{itemize}
On the 1{,}128 papers whose numbering is unambiguous, the results are nearly the same (\cref{app:forward}).

\paragraph{Profile and judge do not rely on memory of the papers.} A judge, or the language models behind the surprisal features, may find the main results of a paper because they have read it beforehand, a form of data contamination \citep{golchin2024time}. I therefore score both on 1{,}418 papers posted from November 2025 to August 2026, after the release of these models (\cref{app:forward}).  Both do \emph{better} on these papers than on the 2010--2018 papers of the same categories. The judge gains $+0.027$ $[+0.013, +0.039]$ and the profile $+0.064$ $[+0.049, +0.080]$ (\cref{app:forward}). Weights fitted on the older papers and applied unchanged to the recent ones do as well ($-0.003$ $[-0.011, +0.004]$), so the profile learned from the past anticipates the authors' choices in papers posted later. Its gain thus comes from the recent papers themselves. \Cref{app:forward} explains this gain by the incompleteness of the older sources. The judge still leads, by $+0.112$ against $+0.148$ before. 

\paragraph{Answer to question (iv).} Selections can be anticipated, by the profile and by the judge. Both rank highly the theorems that curators add years later, both anticipate, more weakly, the results that later papers cite, and both do better on papers posted after the language models they rely on were released.
\section{What lies beyond the current features?}
\label{sec:limits}

I address question (v) of \cref{sec:framework}, what lies beyond the current features, that is, which aspects of interestingness the features of \cref{sec:data} cannot capture. I look for such aspects in three harder settings: definitions, close variants of a theorem, and the distance between the subfields a result connects (\cref{app:limits}).

\paragraph{Definitions are valued for their use.} \mathlib{}'s module documentation also lists
\emph{main definitions}, 2{,}735 of the 43{,}306 definitions written by humans. I compute the same features on them (\cref{app:limits}). The ex-ante features reach only 0.609
$[0.590, 0.631]$, the usage features 0.677, and the judge 0.549. Weights learned on theorems barely transfer to definitions (0.522), and usage dominates for definitions (\cref{app:limits}). A test on \mathlib{}'s history confirms that the form of a definition does not anticipate its use (\cref{app:limits}). Theorems are thus valued for
what they contain and definitions for what they enable, and only the value of definitions needs
hindsight.

\paragraph{The profile finds which result matters, but not which variant.}
\label{sec:rediscovery}
I test whether a profile of ex-ante features can recognize the result that humans singled out among close variants of it. For each of 1{,}467 results singled out by a signaller, a language model writes ``dull'' variants of four types, namely specializations, weakenings, generalizations and trivial variants. Each result and its proved variants form a group, and I fit weights on pairs of a real result and a variant built from other results (\cref{app:pipeline}). Once trivial variants are removed, they rank the real result first in 62\% $[59\%, 65\%]$ of groups, against 45\% for the judge and 26\% for a random choice.

This success, however, comes from one kind of feature, the surprisal of the statement given the text that precedes it in the file. The text of a file before a result is written leading up to it, so these features partly encode hindsight, which is a form of leakage \citep{kaufman2012leakage}. Comparing the same results with lemmas of the same module confirms this leak, and establishes that, without it, the profile still tells a result that humans singled out from an ordinary lemma (\cref{app:limits}). Against variants, I test the profile without the context features, since they carry this leak. The profile then ranks the real result first in only 33\% of groups, against 21\% for a random choice. This remaining gain does not generalize. Weights fitted without one type of variant are close to chance on that type, and prefer the variant for specializations, so they learn how the language model builds each type of variant and not what makes the real result preferable (\cref{app:limits}).

\paragraph{Answer to question (v).} What lies beyond the current features differs for theorems and for definitions. For theorems, the current features capture much of what can be seen when a result is written, and their limit lies in the choice among close variants of a theorem, which only the text written around the theorem that was kept, with hindsight, tells apart. The distance between the subfields a result connects, as I measure it, adds no information beyond the number of these subfields (\cref{app:limits}). For definitions, the ex-ante features predict little, their later use predicts more, and their form does not anticipate this use. Predicting the value of a definition before it is used would require either combining the natural features of definitions more exhaustively, or extracting features more systematically.
\section{Related work}
\label{sec:related}

\paragraph{Interestingness in exploring agents.} Open-endedness, the ability of a system to keep producing artefacts that are new and learnable for an observer \citep{hughes2024openended}, requires a notion of what this observer finds interesting, which an agent that explores without a fixed goal must apply by itself. Intrinsic motivation and novelty search define interestingness relative to the agent and its history \citep{schmidhuber2010creativity,lehman2011novelty}, and other agents query a foundation model, used as a model of human notions of interestingness, to choose the problems they learn to solve next \citep{zhang2024omni,faldor2025omniepic} or to judge their own research ideas \citep{lu2024aiscientist} (\cref{app:related}). The agents above either define interestingness relative to the agent or take it from a foundation model, without comparing it against what people actually select. In this paper, I compare an LLM judge and computable features with the selections made by mathematicians, and find that the two complement each other.

\paragraph{Interestingness in automated mathematics.} Since the 1970s, systems that automate mathematical discovery have also implemented some form of interestingness. They range from the hand-written heuristics of AM and EURISKO \citep{lenat1976am,lenat1984am} to the criteria of Graffiti and HR \citep{fajtlowicz1988graffiti,colton2000interestingness}. Recent systems mostly rely on single criteria, such as the proof/statement ratio \citep{patel2026interesting}, being barely provable \citep{poesia2024minimo,dong2025stp}, or the verdict of critic agents \citep{long2026meca,chen2026moonshine}, while \citet{tsoukalas2025learning} search for a measure, still judged by a single criterion, the production of concepts (\cref{app:related}). My work differs from these systems in two respects. Most of them rely on a single criterion, whereas I combine several criteria, including some of theirs, as features of one profile. Like \citet{tsoukalas2025learning}, I learn how much each aspect matters, but from the selections of mathematicians, with hand-chosen features whose weights can be read (\cref{app:related}).

\paragraph{Other studies of interestingness.} Other works characterize interestingness in mathematics through compression \citep{bengio2024aimathematician,aksenov2026compression,barkeshli2026structure}, or compare human and model judgments of it \citep{matterofinterest2025,gu2024scimuse,quaresma2026geometric} (\cref{app:related}).
\section{Discussion}
\label{sec:discussion}

\paragraph{Discourse centrality is a reasonable but imperfect proxy.} I measure interestingness through what evaluators single out, which records decisions that mathematicians actually took, at scale and independently of my study (\cref{sec:framework}). This proxy remains imperfect. Each label inherits the conventions of its signaller, centrality and interestingness can come apart, for instance for pedagogical reasons, and the labels do not record the reasons for a choice. Direct expert ratings of matched pairs \citep{quaresma2026geometric} would complement these traces (\cref{app:framework}).

\paragraph{Using interestingness in an explorer.} An explorer could use the shared weights of a profile fitted on one library by default, adapt them with a small deviation to the community or person it works for, and combine them with a judge where dependencies are implicit. Two types of decisions need more: choosing among close variants of a theorem, which calls for comparing the variants with each other, and judging a new definition, which an explorer could keep provisionally and judge by its use (\cref{app:limits}).

\section*{Acknowledgements}
This work was supported by the project ``Biography of Fake News with a Touch of AI: Dangerous Phenomenon through the Prism of Modern Human Sciences'' (Project No. CZ.02.01.01/00/23\_025/0008724) via the Operational Programme Jan \'{A}mos Komensk\'{y}. I acknowledge the EuroHPC Joint Undertaking for awarding the project ID EHPC-BEN-2026B07-121, ``Learning the causal structure of two-dimensional subshifts of finite type at scale'' (PI Jan H\r{u}la, Czech Technical University), access to the LEONARDO supercomputer, hosted by CINECA (Italy), through the EuroHPC JU Benchmark Access call.

\section*{AI Usage Statement}
AI tools were used throughout the research and writing process as collaborative assistants. The main research ideas and conceptual motivation originated from me. AI was subsequently used to help operationalize these ideas, develop and execute parts of the experimental workflow, and draft portions of the manuscript.

In particular, Claude Code was used iteratively to assist with implementation and experimentation. I continuously reviewed, discussed and evaluated the code and the proposed approaches, providing feedback and refining the direction of the work throughout this process. The use of AI thus involved an iterative feedback loop between me and the AI system instead of autonomous generation of the research.

AI tools were also used to assist in drafting sections of the paper. I repeatedly reviewed and revised these drafts, corrected inaccurate or misleading statements, requested modifications, incorporated additional ideas, and determined the final content and interpretation. Code, experimental procedures, results, mathematical statements and citations suggested or generated with the assistance of AI were checked by me before being incorporated into the work.

I am responsible for the final methodology, claims, analysis, interpretation and presentation of the work.

\section*{Impact Statement}
This paper aims to advance machine learning for mathematics by understanding what makes mathematical results interesting to human mathematicians, so that discovery systems can select what they produce accordingly. I see no direct negative societal
consequences beyond those general to automated research tools.

\newpage
\appendix
\crefalias{section}{appendix}
\onecolumn
\section{Framework details}
\label{app:framework}

This appendix provides the full motivation of the paper and completes the framework of \cref{sec:framework}. It first explains why interestingness matters for an exploring agent and what I contribute. It then details value functions and shared weights, explains the five questions that I formulate in \cref{sec:framework} and address in the rest of the paper, discusses discourse centrality as a proxy for interestingness, and describes how the weights are estimated and how I chose the evaluators.

\subsection{Motivation and contributions}

\paragraph{Exploring beyond a fixed framework.} Recent contributions of language models to mathematics were produced within a relatively fixed conceptual framework. The definitions, the objects and the
question are fixed in advance, and the model assembles existing pieces, sometimes in ways no one
had thought of. For instance, the unit distance counterexample rests on ideas from class field towers that, in
hindsight, were all in the literature \citep{alon2026unitdistance}. Language models have mastered
induction and are fast mastering deduction, but they have not mastered the abductive leap that produces new meaningful premises
\citep{zahavy2026jump}.

\paragraph{Two views of interestingness.} For an agent to explore mathematics open-endedly and produce something of value, it must know what humans find interesting. The reason lies in what mathematics records. Its purpose is
not to record everything that is true. A machine could enumerate every derivation from the axioms
and list true statements one by one; this was the first baseline of automated reasoning, the
``British Museum algorithm'' \citep{newell1957logic}, and it yields almost nothing anyone would
call mathematics. To create, in Poincar\'e's words, is precisely not to make useless combinations
\citep{poincare1908science}. What mathematics records instead is a minimal conceptual
framework, with a battery of results and proofs, that provides the means to decide efficiently
whether a general statement is true. Existing mathematics can be rebuilt from it, and ideally
anyone with enough time could understand it. This framework grows by focusing on the problems that
its current state cannot reach, and the path towards harder results
cannot be shortcut: each step needs the concepts and results of the ones before. An AI may walk
this path much faster than I do, but its time is limited too, so it must also choose what to
keep. In this
view, which is that of pure mathematics, a result is interesting if it earns its place in that
framework.

On the other hand, pure mathematics is only one point of view. Scientists who need mathematics for their own purposes value it for
what it lets them model, compute or predict. Interestingness is therefore subjective in that
sense, since it depends on who develops or uses the mathematical objects. The divide between pure
and applied mathematics is its starkest form. \citet{hardy1940apology} prized mathematics for
having no use, while \citet{wigner1960effectiveness} marvelled at how useful it turns out to be.
How much of interestingness is shared, and how much belongs to who judges, is one of the questions
I study.

\paragraph{Counterexamples to single proxies.} Most current systems select what they produce with a single proxy, such as the ratio of proof length to statement length, compression or the verdict of a language model (\cref{app:related}). Each single proxy has obvious counterexamples when used alone. A routine computation can
need a long proof for a short statement, while Euclid's proof that there are infinitely many primes
takes a few lines. What is hard for a given prover says as much about the prover as about the
result. A \emph{reformulation} that adds no new fact but shortens many proofs has no novelty and
high value; a \emph{dull generalization} is new, but its proof is a routine adaptation.

\paragraph{Why this matters.} Interestingness is a fuzzy notion, and an agent can pursue it reliably only once I understand
what it depends on, including who evaluates and what can be read from a result alone. This paper is a step towards that understanding, both conceptual
and empirical.

\paragraph{Scope.} I study the question in mathematics, where it can be measured. I
focus on theorems, and test in \cref{sec:limits} whether what I learn on them extends to
definitions. I investigate whether what mathematicians find
interesting can be learned before anyone uses a result. This matters because an agent must decide
what to pursue at the moment it produces a result, without waiting for others to evaluate it.
Mathematics does rely in part on such delayed evaluation, since a conjecture often attracts
attention only after it has resisted attempts for years. Yet someone first chose to state and share
it, on the basis of what was already known, such as examples, partial evidence and its connections
to other results, and this initial judgment is the one an agent has to make. I also investigate
what interestingness consists of. In particular,
I measure how much of it is shared among evaluators and how much depends on the evaluator, and
what a language-model judge knows beyond the structure of a result.

\paragraph{The first two contributions.} \emph{A conceptual framework.} I define precisely the notions
    of evaluator, of the profile of a theorem and of the split of interestingness into a shared and
    a subjective part, and I formulate features drawn from mathematical practice, of which existing
    proxies are special cases. The framework turns vague questions about interestingness into
    testable ones, including how objective its shared part is.

\emph{A measurement of revealed preferences.} Instead of
    surveying mathematicians, I observe what they single out in formal libraries and in papers,
    which I call \emph{discourse centrality}. Unlike the proxies of the literature, computed from
    the result or by a model, it records the community's own judgment, and I argue that it is a
    reasonable proxy for interestingness, against which the others can be tested as features.

\paragraph{Main findings.} The introduction summarizes my answers briefly, and I state them here in full.

    \begin{itemize}
      \item \emph{Interestingness can be learned ex ante} (\cref{sec:prediction}). In \mathlib{}, the
        features predict what every evaluator singles out as well as a language-model judge does, and the
        two are complementary. Later usage adds nothing to the features.
      \item \emph{It is largely shared, and subfields differ little from each other} (\cref{sec:core,sec:libraries}). One set
        of weights carries most of the signal of every evaluator and transfers in part to other libraries.
        It favours a proof that is long relative to its statement, explicit hypotheses, and a deep chain of results drawing on several subfields. Subfields have no weights of their own, while practices and
        individual authors do.
      \item \emph{The judge knows the canon} (\cref{sec:prediction,sec:forward}).
        Its advantage is confined to curated lists, and neither the judge nor the profile relies on memory of the papers, since both do better on papers posted after their training.
      \item \emph{Theorems and definitions are valued differently} (\cref{sec:limits}). Usage
        adds no information for theorems but dominates for definitions, which are valued for what they
        enable.
    \end{itemize}

\subsection{Value functions and shared weights}

\paragraph{Value functions.} The value function of an evaluator $c$ is $\interest_c(x) = f\big(\weights_c, \profile(x)\big)$, as in \eqref{eq:value}, where the notation is as follows.
\begin{itemize}
  \item $\interest_c(x)$ is the interestingness of $x$ for $c$, a number. Only the order it
    induces matters, in the sense that $c$ prefers $x$ to $y$ when $\interest_c(x) > \interest_c(y)$.
  \item $\weights_c \in \mathbb{R}^K$ says how much $c$ cares about each aspect. It contains one
    weight per feature, which is positive if $c$ favours results that score high on it, negative if
    the opposite, and near zero if $c$ ignores it.
  \item $f : \mathbb{R}^K \times \mathbb{R}^K \to \mathbb{R}$ combines the weights of $c$ with the
    profile of $x$ into a value.
\end{itemize}

In this paper I assume that each value function is linear in the features. Any
linear function of $\profile(x)$ can be written $\langle \weights, \profile(x) \rangle$ for a unique vector
$\weights \in \mathbb{R}^K$, so I can take this vector as the weights of $c$, and
$\interest_c(x) = \langle \weights_c, \profile(x) \rangle$. I use within-subfield percentiles as features, meaning that each raw measurement (proof size, dependency height, \ldots) is replaced by its
percentile among the results of $\mathcal{L}$ in the same subfield as $x$. Such a percentile is still a
feature in the sense above, since it is computed from $x$ and $\mathcal{L}$ alone.

\paragraph{Normalization within subfields.} Normalizing within each subfield matters because subfields differ significantly. Proofs may be
longer in some subfields than in others, and some subfields have more main results than others.
With raw values, a feature could then predict which results are singled out only because it
reveals which subfield a result belongs to. A percentile compares a result only with the results
of its own subfield, so it measures how much the result stands out there.

\paragraph{Kinds of evaluators.} Most evaluators are signallers, groups of people defined by a practice through which they single out results. Examples are formalizers marking
the main results of a file, curators compiling lists, the wider community attaching a name to a result (such as ``the enormous theorem'' for the
classification of finite simple groups),
authors choosing what to state in their introduction and later authors citing a result. The same person may
belong to several signallers, since a signaller is defined by its practice. Subfields are evaluators too, since the habits of an area may shape what its members value. An algebraist may, for instance, value most a result that reveals a structure,
such as a classification, and a probabilist a sharp estimate or a limit theorem. Whether subfields
actually value different aspects is one of the questions I formulate (\cref{sec:core}).

\paragraph{Normalized weights.} Since only the order induced by $\interest_c$ matters, $\weights_c$ is determined at best up to a positive factor, so I first normalize each weight vector so that its Euclidean norm is 1, before computing the center $\weights_0$, the mean of the weights $\weights_c$ of the family $\mathcal{C}$, and the deviations $\delta_c = \weights_c - \weights_0$ \eqref{eq:decomposition}. The center $\weights_0$ is the unique vector that minimizes the Euclidean norm of the vector
$\delta = (\delta_c)_{c \in \mathcal{C}}$ that collects all deviations.

\paragraph{An example of shared weights.} Splitting each evaluator's weights into a shared part and a deviation is in line with models of annotators who legitimately disagree \citep{plank2022variation,davani2022disagreement} and with pluralistic alignment \citep{sorensen2024pluralistic}. Take the three signallers of \mathlib{}, namely the formalizers who mark main results, the
curators of lists and the community that names results. If I predict the curators' choices with the center
$\weights_0$ instead of their own weights, how much do I lose? If almost nothing, the three value
results in nearly the same way; if a lot, each has its own notion and the center is a mere
compromise. For instance, if every evaluator rewards a long
proof behind a short statement, that preference is shared; if only one rewards explicit
hypotheses, it lies in that evaluator's deviation.

Recall that labels are binary. For instance, the label of \mathlib{}'s formalizers is 1 for
the theorems a file's documentation lists among its main results, and 0 for all others; the label
of an arXiv author is 1 for the results stated in the paper's introduction
(\cref{sec:data} lists all labels).

\paragraph{Objectivity.} Shared interestingness is not yet objective. The three signallers above are trained mathematicians
of the same culture, with the same education and the same canon. Their agreement may reflect
something about the results themselves, or only what they have learned to value together, and
agreement within one group cannot tell the two apart. Such agreement is evidence of
\emph{intersubjectivity} only. What separates them is that a shared culture tends to break down for evaluators who
do not share it, while objective interestingness should not. I therefore widen the family progressively,
from the signallers of one library (\cref{sec:core}) to subfields of mathematics, to other libraries
built by other communities and to informal mathematics (\cref{sec:libraries}), and finally to
evaluators at later times (\cref{sec:forward}). Shared interestingness is \emph{more objective} the further it holds along this sequence.
Objectivity itself is out of reach, since every family I can test is made of humans doing
mathematics today, and what I can measure is how far the agreement extends.

\subsection{The questions in detail}

\Cref{sec:framework} states five questions briefly. Each of them comes from an intuitive idea about interestingness, which I state here in full.
\begin{itemize}
  \item[(i)] \emph{Is interestingness visible in the result and what it builds on?}
    (\cref{sec:prediction}). By this I mean its statement, its proof and the dependencies below
    it. If it is, the profile should predict discourse centrality, which I test on the signallers of
    \mathlib{}. It should do
    so \emph{ex ante}, that is, from features that can be computed as soon as the result is stated and
    proved, since an exploring agent must judge a result when it produces it, before anyone has used it. I also
    compare the profile with single features, to test whether interestingness is one quantity, and
    with a language-model judge, the main alternative in practice.
  \item[(ii)] \emph{How much of interestingness is shared?} (\cref{sec:core}). If the share $s_c$ is close to 1, an
    agent can learn one notion of interestingness instead of one per evaluator. Where the
    deviations lie matters too. One might expect subfields to value different
    things, and \citet{gowers2000cultures} contrasts the problem-solving culture of combinatorics
    with the theory-building culture of other areas. The deviations could also lie with practices that serve different purposes, such as curating a
    canon or summarizing a file, or with individual taste. Each answer tells an agent whom it has
    to model.
  \item[(iii)] \emph{How objective is shared interestingness?} (\cref{sec:libraries}).
    Agreement among the signallers of \mathlib{} may only reflect the shared culture of one
    community. I therefore test the weights on evaluators from other communities, in two other
    proof assistants and in informal mathematics on arXiv. The further the shared weights
    hold, the more objective they are in the sense above. The answer also determines whether an agent trained in one library
    can be used in another.
  \item[(iv)] \emph{Can a model trained on past selections anticipate later ones?} (\cref{sec:forward}). The tests of question (i)
    predict labels that already exist, many of them set long after the result was proved.
    Hindsight can then help. A language-model judge, and the language models behind
    some of my features, may have read about famous results. A result that became famous may also
    have been restated or reorganized in today's library because
    of its importance. A model could thus
    recognize the reputation a result has acquired, which an exploring agent cannot do with its own
    results. I therefore test whether a model trained on the selections known at a given date
    predicts the selections made afterwards.
  \item[(v)] \emph{What lies beyond the current features?} (\cref{sec:limits}). An exploring
    agent does more than rank finished theorems. I test the profile in situations it faces
    where prediction from a profile may fail.
    \emph{Definitions.} An agent proposes new concepts. A definition seems valued for the theorems it
    makes possible, and these theorems do not exist yet when the definition is proposed. A profile
    computed at that time sees only the form of the definition, and may miss its value.
    \emph{Formulations.} An agent chooses how to state a result among close variants of one theorem,
    such as a special case, a weaker version or a generalization. The variants share most of their
    structure, so their profiles are close, and the profile may not tell which one evaluators
    prefer.
    \emph{Distance.} The results valued most are often those that connect areas far apart, which few
    people master together. The profile measures how many areas a proof draws on, but not how far
    apart they are, so it may miss what makes such a connection valuable.
\end{itemize}

\subsection{Discourse centrality as a proxy}

\paragraph{What is observed.} Of the two ingredients of
the value function \eqref{eq:value}, the profile is observed, since I compute the features myself, but the
weights $\weights_c$, and hence the interestingness $\interest_c(x)$, are not, since a formalizer or an
author does not report a number for each result.

\paragraph{Labels as data.} Labels are the data an evaluator
leaves, features are the inputs from which I predict them, and the weights $\weights_c$ are what
the prediction recovers, as when preferences are learned from clicks or comparisons \citep{joachims2002optimizing,christiano2017deep}. Centrality and interestingness are different functions, since $y_c$ takes values in $\{0,1\}$ and
$\interest_c$ in $\mathbb{R}$, so the first cannot stand in for the second value by value.

\paragraph{Discourse and dependency centrality.} Discourse centrality should not be confused with \emph{dependency centrality}, how
central a result is in the dependency graph, that is, how many other results use it in their
proofs. The first is what people say about a result, and the second is what proofs do with it.
Dependency centrality is a feature, available only after a result has been used, whereas
discourse centrality is what I try to predict from the features.

\paragraph{Estimating a probability from single observations.} Under monotone selection \eqref{eq:monotone}, the probability that $c$ singles out $x$ increases with $\interest_c(x)$, and the label is a coarse and noisy trace of the value function. Each result is observed once, as singled out or not, so the probability
is not estimated result by result. Since $\interest_c(x)$ depends on $x$ only through its profile,
results with similar profiles share the same probability, and the fraction of them that $c$ singled
out estimates it, much as one estimates the bias of many coins of the same make by tossing each
once. The noise stands for everything that decides a selection beyond the profile.

\paragraph{Why monotone selection is reasonable.} The assumption of monotone selection is reasonable because singling out is what the value function is for. Mathematicians
put forward, name, teach and cite the results they find interesting, and leave in the background
the auxiliary lemmas they do not; a result that no one ever singles out is hard to call
interesting for anyone. The labels also come from independent practices, so a pattern common to
all of them is unlikely to be an artefact of one. However, this proxy is imperfect.

\paragraph{Where centrality and interestingness come apart.} Centrality also depends on factors that do not make a result more interesting to its evaluator.
A result by a well-known author is cited, named and listed more often than an equally interesting
result by a less visible one, so it is more central without being more interesting. Priority acts
in the same way. A name goes to whoever first stated or proved a result, and a later result that
evaluators find just as interesting, such as a cleaner or more general version, may remain less
central. A curriculum may include a result because it is easy to state and prove in class, which
makes it central for pedagogical reasons. Finally, a technical lemma that many papers cite because
their proofs need it is singled out by citing authors without being found interesting. In such
cases $y_c$ depends on more than $\interest_c$, and \eqref{eq:monotone} holds only approximately.
The reverse gap also exists, with results that are interesting without being central. A result
stated in a little-read paper, or buried as a lemma in a long proof, can be overlooked by
evaluators who would have found it interesting. A result can also be recognized only later, when
another development reveals its value, and it is then interesting for later evaluators while it is
not yet central for current ones.

\paragraph{Trends and history.} It may be objected that centrality also depends on intellectual trends and on the history of
mathematics. These factors, however, act on centrality and on interestingness in the same way,
because they act on centrality only through interestingness. A theorem that settles a problem
left open for decades is singled out because the long wait makes its solution more valuable to
mathematicians, and results in an area that attracts attention are singled out because the
community values that area at that time. In both cases the factor first changes how much the
evaluator values the result, and the evaluator then singles out what it values. The visibility of
authors or priority, by contrast, change what is singled out without changing how much the result
is valued. Since my framework makes interestingness relative to an evaluator, including its time,
trends and history are part of what I measure, and they leave \eqref{eq:monotone} intact.

\paragraph{Why discourse centrality remains a reasonable proxy.} First, the
distortions above act on different signallers. The visibility of authors weighs on citations and names,
priority on names, pedagogy on curricula, and none of them decides what the formalizers of a file
list as its main results. A pattern found in every signaller is therefore unlikely to come from
one distortion. Second, several distortions are invisible in a result. The identity of its authors
and the date of its discovery appear in no feature, and names are hidden from the language
models. These distortions add noise to the labels. If this noise occurs independently of the features, the AUROCs I report, which are computed against these labels, understate how well the profile ranks results by interestingness, but the noise does not change which results the weights favour. It biases the weights only where it correlates with a feature, for instance if results central for pedagogical reasons tend to have short statements. Third, the main alternative is weaker. Surveying mathematicians would yield stated preferences, on small samples and in
answer to a question I would choose. Discourse centrality records decisions that
mathematicians actually took, at scale and independently of my study. Several tests also
target specific distortions, such as the reputation of a result and documentation (\cref{sec:prediction}) and the memory
of a language model (\cref{sec:forward}).

\paragraph{Conventions of signallers.} Each label inherits the conventions of its signaller, such as the documentation habits
of \mathlib{}, the purpose of a curated list, the history of naming or the way authors frame their
papers. Part of the differences between signallers may therefore come from these conventions, and
an author's deviation may partly concern presentation, since \Lintro{} records what the author
chooses to put first, although its most stable parts concern content (\cref{app:core}).

\paragraph{What the labels do not record.} The labels record which results are singled out, without the
reasons for this choice. Direct expert ratings of matched pairs, as in the survey of
\citet{quaresma2026geometric}, would complement these traces with judgments given for this purpose,
and would separate taste from presentation.

\paragraph{Interestingness as practised.} What I measure can thus be considered as interestingness as practised. It inherits the habits and
blind spots of each evaluator, and I read every result with that in mind. The labels overlap little, so I report them separately instead of
combining them into one score.

\subsection{Estimation}

I estimate each $\weights_c$ by logistic regression on the labels of $c$. Selections are noisy, since an evaluator overlooks some results, as in learning from positive and unlabeled data \citep{elkan2008learning}, and singles out others for reasons
outside the profile. Fitting $\weights_c$ inverts
\eqref{eq:monotone}: it finds the weights whose score best separates the results $c$ singled out
from the others, and the fitted score is my estimate of the value function of $c$. In \cref{sec:core} I fit $\weights_0$ and all $\delta_c$ jointly, penalizing them by
$\lambda \big(\|\weights_0\|^2 + \sum_c \|\delta_c\|^2\big)$ (ridge), as in regularized multi-task learning \citep{evgeniou2004regularized,daume2007frustratingly}. Since $\weights_0$ is penalized too, at the optimum the deviations sum to $\weights_0$, so the
fitted $\weights_0$ is $|\mathcal{C}|/(|\mathcal{C}|+1)$ times the mean of the fitted $\weights_c$
(three quarters of it for three signallers). It is thus the mean of \eqref{eq:decomposition}
multiplied by a constant, which changes neither the signs of the weights nor any AUROC. The penalty
$\lambda$ keeps each $\delta_c$ small unless the data require it. Since only the order induced by $\interest_c$ matters, I score it by $\mathrm{AUROC}_c$.

\subsection{Why I chose these evaluators}

I work with a finite set of evaluators, known only
through the labels they leave. The pool of all evaluators of mathematics, every mathematician and
every user of mathematics, past and present, is not a population I can sample from, so I make no
claim of representativeness in the statistical sense. I choose evaluators by three criteria.
\begin{itemize}
  \item \emph{They speak for their community.} Each is a party whose role is to decide what is
    central. \mathlib{}'s module documentation is approved in review by the library's maintainers,
    curated lists and names are kept or conferred by a community, and authors decide what their
    papers put forward. Their labels are thus the community's own record of what is central.
  \item \emph{Their labels are recorded at scale.} Each label covers a whole library or thousands
    of papers, enough to fit and test weights. The labels were also produced independently of
    my study, so they cannot reflect what I expected to find.
  \item \emph{They differ} along axes on which a notion of interestingness could plausibly differ,
    so that agreement among them is informative. They differ in \emph{scope} (one file or a list of
    famous theorems), in \emph{time} (from when a result is written to decades later), in
    \emph{who} decides (one author, a few curators or a diffuse community) and in \emph{medium}
    (three proof assistants and informal papers).
\end{itemize}
This choice has limits. First, those who decide what is central are not the whole community.
Reviewers, curators and authors of introductions may be more conservative than the community they
speak for, and favour the canon, the results of their own area, or what is easy to present.
Second, the family has known biases. \mathlib{} covers mostly undergraduate and classical graduate
mathematics, and its formalizers are a particular community. This is changing fast, since formal proofs
now exist of Fermat's Last Theorem \citep{anthropic2026flt} and, built on \mathlib{} though
outside it, of the classification of finite simple groups \citep{axiom2026cfsg}, so libraries
covering large parts of research mathematics may soon allow my analysis to be repeated on them.
Curated lists and names favour classical results; my arXiv sample is from 2010--2018 and mostly
pure mathematics; applied mathematics, and the scientists who use mathematics, are not
represented. Third, and most fundamentally, what these evaluators record is the interestingness of
a community, with its history and its arbitrary choices, and no absolute
interestingness independent of any community can be read from them. My claims about shared interestingness are claims about this family,
and \cref{sec:libraries,sec:forward} widen it as far as the data allow.

\section{\mathlib{}: extraction, labels, features and judge}
\label{app:mathlib}

This appendix describes how I build the \mathlib{} data of \cref{sec:data} and provides the details behind \cref{sec:prediction}: the extraction of declarations and labels, the definition of each feature, the evaluation, the language-model judge, further results on statement, proof and usage features, the controls for shortcuts, and the features I computed but do not use.

\subsection{Extraction and labels}
\label{app:extraction}

\paragraph{The corpus.} From \mathlib{} v4.34.1 I extract every declaration together with its statement,  its proof, the declarations its proof uses directly, the module it belongs to, the
documentation at the top of that module, and its position in the file. I keep only the hand-written theorems, 237{,}115 of the 280{,}657 theorems of this version. I exclude the auxiliary declarations that Lean generates (listed below, 24{,}679 theorems), the theorems of infrastructure directories (2{,}955) and instances (17{,}254). The additive versions that the attribute \texttt{to\_additive} generates from multiplicative lemmas have the source position of the lemma they come from, so they are kept. They rarely matter for the labels, since few positives share their source position with another theorem (at 63 source positions for the 3{,}205 positives of \Lmain{}, 3 for \Llists{} and 2 for \Lnamed{}), and the other theorem is then nearly always a negative.

\paragraph{Declarations.} A Lean 4 metaprogram run on \mathlib{} v4.34.1 exports, for every
declaration: name, kind, module, source range, docstring, statement and proof term with their
tree size, DAG size and depth, binder counts by kind (explicit, implicit, instance, hypotheses),
and the constants used in the statement and in the proof. An edge $x \to y$ means that $x$ uses
$y$ in its statement or proof. Auto-generated declarations (equation lemmas, \texttt{match} and
\texttt{proof} auxiliaries, injectivity and \texttt{sizeOf} lemmas) and declarations without a
source range are excluded, as are infrastructure directories (\texttt{Tactic}, \texttt{Lean},
\texttt{Util}, \texttt{Testing}, \texttt{Deprecated}, \texttt{Control}, \texttt{Init},
\texttt{Std}). A declaration's subfield is the first component of its module after
\texttt{Mathlib}.

\paragraph{\Lmain.} Module docstrings (\texttt{/-! \ldots -/}) are parsed for second- or third-level
headers named \emph{Main results}, \emph{Main result(s)}, \emph{Main statement(s)} or \emph{Main
theorem(s)}. Backticked identifiers in these sections are resolved against the declaration
table. A mention counts if it leads its list item, or if it resolves unambiguously to a
declaration of the documented module. Sections on main \emph{definitions} provide the label for
definitions (\cref{sec:limits}).

\paragraph{\Llists.} A theorem is positive for \Llists{} if it is linked from one of four lists maintained alongside \mathlib{}, namely Wiedijk's \emph{100 theorems}, a list of famous theorems used to compare proof
    assistants; its extension to \emph{1000+ theorems}, drawn from Wikipedia; an overview of the
    library's content by area; and the topics of a standard undergraduate curriculum (414
    positives). These are the entries of \texttt{docs/100.yaml}, \texttt{docs/1000.yaml},
\texttt{docs/overview.yaml} and \texttt{docs/undergrad.yaml} that resolve to declarations.

\paragraph{\Lnamed.} A declaration is named if its own docstring mentions a named result: up to four
capitalized words (allowing hyphenated names such as Cauchy--Schwarz and an optional possessive),
up to two lower-case modifiers, then a result word (theorem, lemma, inequality, formula, identity,
criterion, conjecture, principle, law, reciprocity, duality, correspondence, decomposition,
classification). Sentence-initial words (``The'', ``This'', ``Main'', \ldots) are not names.
This pattern also matches descriptive phrases (``Auxiliary lemma'', ``Key technical lemma''), and only
49\% of positives carry a possessive or hyphenated name. The \emph{eponymous} variant keeps
positives whose docstring mentions a proper name (possessive, hyphenated, or a capitalized word
that \mathlib{}'s docstrings almost never write in lower case), 1{,}098 of 1{,}684. On it, ex-ante
AUROC is 0.836 $[0.801, 0.869]$, usage adds $+0.001$, pooled minus own-subfield weights is $+0.013$,
and matched minus mismatched is $+0.057$ (vs.\ \Lmain) and $+0.010$ (vs.\ \Llists). An apparent border
effect on the original label ($+0.062$ $[+0.029, +0.082]$ for own weights in Probability,
Combinatorics and Computability) disappears on it ($-0.019$ $[-0.049, +0.020]$), since it came from
descriptive phrases in border subfields.

\paragraph{Overlap of the labels.} The labels barely overlap: measuring this overlap by the Jaccard index, I obtain 0.03
for \Lmain{} and \Llists{}, 0.05 for \Lmain{} and \Lnamed{}, and 0.10 for \Llists{} and \Lnamed{}.

\subsection{Feature definitions}
\label{app:facets}

\paragraph{Overview.} I group the features by the time at which they become available. \emph{Ex-ante} features can be computed once a result is stated and proved, and \emph{ex-post} features need later usage, which an exploring agent does not have when it produces a result.
\begin{itemize}
  \item \emph{Statement} (ex ante). How long the statement is, and how many arguments it
    takes, by kind: explicit arguments, arguments Lean infers by itself, structural assumptions
    (such as ``$G$ is a group''), and the other hypotheses. How \emph{surprising} the statement is to a
    language model (\emph{Qwen3-8B-Base} \citep{yang2025qwen3}, \emph{DeepSeek-Prover-V2-7B} \citep{ren2025deepseekproverv2}), measured per token, either alone or
    after the text of the file that precedes it. How \emph{novel} it is: one minus its similarity to
    the closest earlier declaration of the library, in the embedding space of \emph{Qwen3-Embedding-0.6B} \citep{zhang2025qwen3embedding}.
  \item \emph{Proof} (ex ante). The height of the dependency tree below the theorem and the
    size of its proof; the ratio of the length of the proof to that of the statement \citep{patel2026interesting}; how many areas it draws on, measured by the entropy \citep{shannon1948mathematical} of the distribution of subfields among the
    declarations it uses (0 if they all come from one subfield, higher the more evenly they spread
    over several); and how novel the
    theorem is with respect to those results.
  \item \emph{Usage} (ex post), or \emph{dependency centrality}: how many later results use
    the theorem in their proofs, its PageRank \citep{page1999pagerank} in the dependency graph, which is high when the result is used by many results that are
    themselves heavily used, how many subfields its users
    come from, and how much the library saves by stating it once, a form of compression gain
    \citep{bengio2024aimathematician}.
\end{itemize}

The definitions below make each feature precise.

All features enter the models as within-subfield percentiles; missing values are mean-imputed. For the weights reported in \cref{sec:core}, each feature is then \emph{standardized}, that is, shifted and rescaled to mean 0 and standard deviation 1 over all theorems, so that a weight is the change in log-odds of being singled out for one standard deviation of its feature, and weights of different features are comparable.

\begin{itemize}
  \item \textbf{Statement size}: tree size of the statement. \textbf{Binders}: number of explicit,
    implicit, instance and hypothesis binders.
  \item \textbf{Proof/statement ratio} \citep{patel2026interesting}: tree size of the proof term
    divided by that of the statement.
  \item \textbf{Dependency height}: length of the longest chain of uses below the result.
    \textbf{Unwrapped size}: log of the size obtained by recursively inlining every dependency.
  \item \textbf{Dependency-subfield entropy}: Let $D(x)$ be the set of declarations that $x$ uses directly, in its statement or its
    proof, and let $p_k$ be the fraction of $D(x)$ that lies in subfield $k$. The feature is the
    Shannon entropy $H(x) = -\sum_k p_k \log_2 p_k$, summed over the subfields with $p_k > 0$, and
    $H(x) = 0$ when $D(x)$ is empty. It is 0 when all dependencies lie in one subfield and
    $\log_2 m$ when they are spread evenly over $m$ subfields. Unlike the number of subfields
    involved, it is barely affected by a few dependencies on basic subfields, such as a single lemma
    from Order in an algebraic proof.
  \item \textbf{Surprisal}: mean negative log-probability per token of the anonymized statement
    (\texttt{theorem thm \ldots}), bare (after the header \texttt{-- Lean 4 (Mathlib)}) or after
    the preceding 1{,}536 tokens of its file (with the result's own docstring removed). Models:
    Qwen3-8B-Base and DeepSeek-Prover-V2-7B.
  \item \textbf{Novelty vs.\ earlier results}: one minus the cosine similarity
    (Qwen3-Embedding-0.6B) to the most similar declaration among the 100 nearest neighbours that
    the result could have been stated after: in a transitively imported module, or earlier in the
    same module. \textbf{Novelty vs.\ dependencies}: one minus the maximum similarity to the
    declarations it uses directly.
  \item \textbf{Usage} (ex-post): number of hand-written users; PageRank on the use graph, where each declaration passes its score, split equally, to the
    declarations it uses, with damping factor 0.85 and declarations that use nothing spreading their
    score uniformly; number
    of distinct subfields among users and share of users outside the result's subfield;
    \textbf{compression gain}, $\max(u_p - 1, 0)\cdot|\text{proof}| - |\text{statement}|$, where
    $u_p$ is the number of proofs using the result, of which I take the signed logarithm. It measures how much the library
    saves by stating the result once. Writing its proof inside each of the $u_p$ proofs that use it
    would cost $u_p$ times the size of its proof, while stating it as a separate theorem costs its
    proof once plus its statement, and the difference is the formula above.
  \item \textbf{LLM judge}: system prompt ``You are an expert research mathematician.''; user
    prompt ``Here is a result from Mathlib, the Lean 4 library of formalized mathematics:
    [statement]. How mathematically interesting is this result: how significant or central would
    working mathematicians consider it? Answer with a single digit from 0 (a routine technical
    lemma) to 9 (a famous landmark theorem). Answer with the digit only.'' The score is
    $\sum_d d\,p(d) / \sum_d p(d)$ over the digit tokens among the top-20 first-token
    log-probabilities.
\end{itemize}

\paragraph{Anonymization.} For the features that use a language model, and for the language-model judge, the names of theorems are replaced by a
placeholder before the model sees the theorem, since a name such as \texttt{fermat\_little} may reveal
how well known a theorem is.

\subsection{Evaluation}

I measure how well a score orders results by the AUROC. The AUROC depends only on the order a score induces and is unaffected by how rare positives are. I compute
it within each subfield, pairing results of the same subfield only, so that a score gains nothing
from knowing which subfields have more main results, and average over the subfields with at least
20 positives. Confidence intervals are 95\%
percentile bootstrap intervals that resample modules \citep{efron1993bootstrap,field2007bootstrapping}, with 300 resamples.

\subsection{The language-model judge}

\emph{Qwen2.5-32B-Instruct} \citep{qwen2024qwen25}, and for robustness
\emph{Llama-3.1-70B-Instruct} \citep{grattafiori2024llama3}, receives the statement of a result, or the whole declaration with
its proof, and is requested to rate how significant working mathematicians would consider it, on a scale from 0 (a
routine technical lemma) to 9 (a famous landmark theorem); its prompt is given with the feature definitions (\cref{app:facets}). Instead of keeping the digit it writes, I average the digits weighted by the probability the model
assigns to each of them, which yields a finer score \citep{liu2023geval}.

I use such a judge for three purposes. It is a
\emph{baseline}, which tests whether a profile of computable features does as well as a large
model with broad knowledge of mathematics. It is a \emph{probe}, since where it does better it knows something the
profile does not, such as which results belong to the canon. And it is an \emph{evaluator} in the
sense of \cref{sec:framework} whose value function, unlike that of any human evaluator, I can
query directly.

\paragraph{The judge shown the proof.} \Cref{tab:e1proof} compares the ex-ante features with the judge shown the whole declaration with its proof, on the subset of about 24{,}000 theorems for which it was shown, and with a larger judge of another family. The features minus the judge shown the proof are $-0.013$
$[-0.029, +0.002]$ on \Lmain{}, $-0.045$ $[-0.091, -0.009]$ on \Llists{} and $+0.018$
$[-0.008, +0.037]$ on \Lnamed{}. The features and this judge thus tie, except on curated lists, where the judge leads. The larger
judge (\emph{Llama-3.1-70B}, statement only) exhibits the same pattern. The two judges agree closely with each other (Spearman correlation 0.895), so the
complementarity of judges and features does not come from a quirk of one model.

\begin{table}[h]
\centering
\caption{Further predictors of \mathlib{} labels (within-subfield AUROC). Top: statement and proof features separately, and ex-ante features with usage, on all theorems. Bottom: the judge shown the proof, on the subset of about 24{,}000 theorems for which the whole declaration was shown.}
\label{tab:e1proof}
\small
\begin{tabular}{@{}lccc@{}}
\toprule
Predictor & \Lmain & \Llists & \Lnamed \\
\midrule
Statement features & 0.674 & 0.800 & 0.793 \\
Proof features & 0.756 & 0.744 & 0.758 \\
Ex-ante + usage & 0.783 & 0.836 & 0.829 \\
\midrule
Ex-ante & 0.748 & 0.775 & 0.812 \\
Judge 32B, statement + proof & 0.761 & 0.820 & 0.794 \\
Judge 70B, statement & 0.723 & 0.829 & 0.804 \\
Ex-ante + judge 32B (proof) & 0.790 & 0.834 & 0.835 \\
\bottomrule
\end{tabular}
\end{table}

\paragraph{Why the judge leads on curated lists.} One
explanation is that it remembers well-known statements. If so, its lead should be largest on the lists
of landmark theorems and vanish on the list of an undergraduate curriculum, and it should also lead
on named results, which are well known by construction. Neither happens. Its lead is the same on every
list ($+0.046$ to $+0.061$ over the features, curriculum included), and it \emph{trails} the
features on named results ($-0.023$ $[-0.042, -0.007]$). What the judge knows is the
\emph{canon}, the set of results that are taught, surveyed and listed as landmarks. This knowledge
concerns the community more than the result, and no profile of the result can contain it.

\subsection{Statement, proof and usage features}

Beyond whether the profile predicts the labels, \cref{sec:prediction} addresses three further
questions. Does any single feature do as well, as single proxies assume? Does later usage help, as
it would if discourse centrality were mostly dependency centrality, that is, if evaluators singled
out the results that many other results use? And how does the profile compare
with a language-model judge? This subsection completes the answers to the first two.

\paragraph{Statement and proof features.} Statement and proof features each predict part of the signal of the three signallers of \mathlib{} (\cref{tab:e1proof}, top), but the evaluators read it in different places. The formalizers' main results are mostly readable from the proof
(adding proof features to statement features gains $+0.108$ $[+0.095, +0.123]$ on \Lmain{}),
whereas results put on lists or given a name are already largely readable from the statement
(adding statement features to proof features gains $+0.089$ $[+0.067, +0.110]$ on \Llists{}). What makes a theorem the main result of its file appears mostly in its proof, in how large
it is and how deep the chain of results below it goes; what gets it listed or named is already visible in how it is stated.

\paragraph{Discourse and dependency centrality.} Later usage adds no information to the ex-ante features (\cref{sec:prediction}). This also provides evidence that
discourse centrality differs from dependency centrality, in line with the observation that
usage counts measure technical utility \citep{mathlibnetwork2026}.

\subsection{Controls for shortcuts}

I rule out three shortcuts \citep{geirhos2020shortcut}. \emph{Position.} I record the position of each theorem in its file, but leave it out of the features
unless stated. Main results tend to sit at the end of a file, a convention of how files are written
that says nothing about the result itself, but position adds at most $+0.01$ once statement and proof features are in (\cref{app:side}).
\emph{Documentation.} Every named result has a docstring, and documented theorems may be special;
restricted to documented theorems, AUROCs drop by 0.05--0.08 but the conclusions hold, with the
features ahead of the judge on \Lmain{} and \Lnamed{} ($+0.043$, $+0.047$) and behind on \Llists{}
($-0.061$). \emph{Context.} Surprisal in the file context could exploit module documentation,
which mentions most main results; it does not, since its AUROC is lower on the theorems the
documentation mentions (0.43) than on the others (0.55).

\subsection{Further features}
\label{app:side}

I also computed two features based on an automated prover (\emph{DeepSeek-Prover-V2-7B}), on a
sample of 8.5k results. The first is how hard the theorem is for the prover, measured by the
fraction of 32 attempts that succeed. The second is how much easier it becomes when the prover is
given, as hints, up to three of the lemmas used in the real proof, chosen among the most specific
ones; a large gain suggests that the proof needs a non-routine ingredient. Both are
mostly redundant with proof size, and I do not use them further. I also report the coverage of the novelty features and the effect of the position in the file.

\paragraph{Prover difficulty.} I ran \emph{DeepSeek-Prover-V2-7B} with 32 attempts per theorem (pass@32) on 8{,}539 theorems. Labelled results are harder to prove. The prover solves at least once 8.2\% of positives, 17.5\% of matched negatives, that is, unlabelled theorems of the same module or subfield with a statement of similar size, and 38.1\% of random theorems. Alone, the feature reaches an AUROC of 0.53 to 0.56, and it adds about $+0.01$ to the features computed from the dependency graph. Its signal is therefore real, but mostly redundant with proof size.

\paragraph{Non-routineness.} Up to three hint lemmas raise the solve rate from 13.2\% to 22.8\%, with similar gains for positives (from 0.073 to 0.159), matched negatives (from 0.134 to 0.236) and random theorems (from 0.307 to 0.413). Since the gain is similar in all three groups, it does not help to predict the labels.

\paragraph{Novelty.} The novelty features are available for 77.8\% of theorems, and the missing ones are mostly generated declarations. Alone, novelty with respect to earlier results reaches an AUROC of 0.57 on \Lmain{}, 0.51 on \Llists{} and 0.47 on \Lnamed{}, and it has a robust positive weight for \Lmain{} in the joint model.

\paragraph{Position in file.} Main results sit at the end of files, so position has a positive
weight for \Lmain{} in every subfield. It adds at most $+0.01$ to sets with statement and proof features, and
$+0.03$ to $+0.08$ to usage-only sets, because position correlates with usage.

\section{The shared part in detail}
\label{app:core}

This appendix provides the details behind \cref{sec:core}: the shared weights and how to interpret them, the transfer between the signallers of \mathlib{} and where they diverge, the comparison of subfields, and the test of individual authors' deviations.

\subsection{The shared weights}

\begin{figure}[h]
\centering
\includegraphics[width=0.5\linewidth]{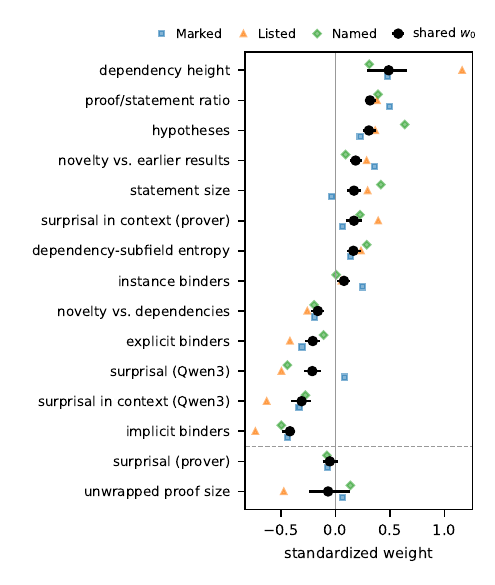}
\caption{Shared weights $\weights_0$ of the three \mathlib{} signallers (black), as fitted by the joint model of \cref{sec:core}, on
standardized ex-ante features, with 95\% intervals from 300 refits on bootstrap samples of modules,
and each signaller's own fitted weights $\weights_c$ (coloured markers). Features above the
dashed line have a shared weight whose interval excludes 0.}
\label{fig:core}
\end{figure}

\Cref{fig:core} shows the shared weights of the joint model of \cref{sec:core}, together with each signaller's own weights.

\paragraph{Interpreting the shared weights.} Each shared weight indicates the direction in which its feature moves the score of a result when the other
features are held fixed. Only the signs of the shared weights can be interpreted with confidence. A shared weight measures the
effect of its feature with the others held fixed, so its size depends on which other features are in
the model. When two features almost always rise and fall together, raising the weight of one and
lowering that of the other barely changes the predictions, so the data cannot tell how their common
effect should be divided between them. This happens for dependency height and the size of the proof
with all its dependencies unfolded (Spearman correlation 0.96): height gets $+0.49$ and unfolded
size an interval around 0, and another sample of modules could shift weight from one to the other.
The same holds for surprisal in context under the two language models (0.76), whose shared weights have
opposite signs. Most other features are only weakly correlated (median absolute correlation 0.11), and their shared weights are more stable.

Even the sign of a shared weight can mislead, since it reflects only the part of its feature that
the other features do not already capture, and it can therefore differ from the effect of the
feature alone. In order to confirm that these signs actually describe each feature taken as a whole, I rank
results by each feature alone, without holding the others fixed. Every feature whose shared weight has an interval on one side of 0 then ranks results in the
direction of that weight, or is neutral (AUROC within 0.01 of $\tfrac12$) for some signallers. The
exception is surprisal in context under the prover's model, which is neutral alone and whose
positive weight only offsets its correlation with the other model (\cref{app:facets}).

\paragraph{Directions.} Single-feature within-subfield AUROC of each ex-ante feature (\Lmain{} /
\Llists{} / \Lnamed{}; they differ by at most 0.006 from those of \cref{tab:e1}), and the
sign of its shared weight in \cref{fig:core} ($+$, $-$, or 0 when the interval contains 0):
dependency height 0.725 / 0.716 / 0.710 ($+$); proof/statement ratio 0.686 / 0.672 / 0.694 ($+$);
hypotheses 0.561 / 0.649 / 0.681 ($+$); novelty vs.\ earlier results 0.554 / 0.506 / 0.505 ($+$);
statement size 0.502 / 0.603 / 0.652 ($+$); surprisal in context (prover) 0.505 / 0.514 / 0.508
($+$); dependency-subfield entropy 0.636 / 0.651 / 0.657 ($+$); instance binders 0.567 / 0.518 /
0.547 ($+$); surprisal (prover) 0.483 / 0.392 / 0.358 (0); unfolded proof size 0.723 / 0.711 /
0.718 (0); novelty vs.\ dependencies 0.441 / 0.420 / 0.410 ($-$); explicit binders 0.448 / 0.456 /
0.489 ($-$); surprisal (Qwen3) 0.491 / 0.298 / 0.295 ($-$); surprisal in context (Qwen3) 0.502 /
0.426 / 0.451 ($-$); implicit binders 0.430 / 0.439 / 0.474 ($-$). Unfolded proof size has no clear
shared weight although it predicts well alone, because it is almost the same feature as dependency
height (Spearman correlation 0.96).

\subsection{Transfer between signallers}

\Cref{tab:transfer} reports, for each pair of signallers, how well weights trained on one label rank the results of another.

\begin{table}[t]
\centering
\caption{Transfer between signallers in \mathlib{}. Each column indicates the label the model was
trained on, each row the label it is evaluated on (within-subfield AUROC). The diagonal, in bold, is
the matched label. The models use the ex-ante and usage features, hence small differences with \cref{tab:e1}, since usage changes nothing (\cref{sec:prediction}).}
\label{tab:transfer}
\small
\begin{tabular}{@{}lccc@{}}
\toprule
Evaluated $\backslash$ trained on & \Lmain & \Llists & \Lnamed \\
\midrule
\Lmain{} (formalizers) & \textbf{0.783} & 0.737 & 0.706 \\
\Llists{} (curated lists) & 0.793 & \textbf{0.836} & 0.818 \\
\Lnamed{} (community) & 0.772 & 0.809 & \textbf{0.829} \\
\bottomrule
\end{tabular}
\end{table}

\paragraph{Documented theorems.} Weights trained on the label of one signaller always rank the results of another worse than weights trained on that label. One artefact could create these gaps. Every named result has a docstring, so the weights for
\Lnamed{} may partly learn what documented theorems look like. However, when I restrict to documented
theorems, the gaps persist ($+0.018$ to $+0.093$).

\paragraph{Where signallers diverge.} The signallers' own weights $\weights_c$ differ most on statement size, dependency height, novelty with respect to
earlier results, surprisal of the statement alone and the number of structural assumptions (such as
``$G$ is a group''). Formalizers (\Lmain{}) single out novel results with structural assumptions
(novelty $+0.36$, structural assumptions $+0.25$) and are indifferent to statement size ($-0.03$).
Curated lists (\Llists{}) favour results deep in the dependency tree (height $+1.16$). The community
(\Lnamed{}) names results with rich statements that look familiar to a language model (statement
size $+0.42$, surprisal $-0.44$).

\subsection{Subfields}

\paragraph{Noise of own-subfield weights.} Weights fitted for each subfield separately do worse than pooled weights (\cref{sec:core}). Own-subfield weights
are fitted on fewer results, so they are noisier, and nothing in the data compensates for that
noise. Likewise, the more freedom subfield deviations are given (the weaker their penalty), the lower
the AUROC (from 0.781 to 0.758 on \Lmain{}).

\paragraph{Border subfields.} Subfields at the border of mathematics, such as
Probability, Combinatorics and Computability, might differ most from the rest. They do not differ
more than central groups, made of subfields from the center of the library with a similar number of
results. For instance, on results named after a person, weights fitted on the border subfield itself
do no better than weights fitted on the other subfields ($-0.019$ $[-0.049, +0.020]$).

\subsection{Individual authors}

\paragraph{The label and the sample.} On arXiv (\cref{sec:libraries}), the
label \Lintro{} records which results the authors of a paper state in its introduction, so each
author is an evaluator. I take the 500 most prolific arXiv authors of 2010--2018 with at least 12 papers (497 of them have at least 8 papers in which some results are stated in the introduction and others are not, with a median of 19 papers and 98k candidate results in total).

\paragraph{Construction of the sample.} I identify an author by their last name and full first name, and keep the authors with at least 12 eligible mathematics papers in 2010--2018 spread over at most four primary categories, which limits merged homonyms. I take the 500 authors with the most papers, with up to 25 papers each (9{,}064 papers, of which 7{,}166 were scored for this study).

\paragraph{A stable taste.} Each author's deviation is fitted from the weights fitted on the papers of all authors together, with an L2 penalty $\alpha \|\delta_c\|^2$. To test whether a deviation
reflects a stable taste or noise, I split each author's papers into two halves and estimate the author's weights $\weights_0 + \delta_c$ on each half by logistic regression, with $\weights_0$ set to the weights fitted on all papers of all authors, the same for both halves, so that each half yields its own estimate of $\delta_c$. I then compute the correlation between the two deviations, seen as vectors with one entry per feature. If the deviation reflects a stable taste, the two halves should yield similar vectors. Averaged over authors, the correlation is $+0.15$ to $+0.27$ (for $\alpha$ from 1 to
100). To judge whether this correlation is high, I compare it with two baselines. For halves taken from two different authors it is $+0.01$,
and for two random sets of papers from the author's main arXiv category, which measures how
consistent a subfield is, it is $+0.04$ to $+0.05$. Authors are thus four to six times more
consistent than their subfield. All these author correlations have $p < 0.001$ (200 permutations). Individual taste is therefore real and stable, and it is small compared to the
shared weights.

\paragraph{Content and presentation.} Their deviations concern content as well as presentation. The most
stable ones measure how much an author's choices follow dependency height (split-half correlation
0.46) and the judge's score (0.41). With a strong penalty, an author's deviation adds $+0.008$
AUROC on their held-out papers.

\paragraph{Purpose and taste.} Part of the differences between the signallers of \mathlib{} may come from what each signaller is meant to contain, beyond simply
signifying that it finds the results it includes interesting. A list of landmark theorems, for instance, is bound to contain deep
results, and a curriculum the results that students must learn. The bootstrap establishes that the
differences are not noise, but it cannot tell whether they come from such a purpose or from what the
signaller values, its taste. The variations between individual authors partly remove this ambiguity. They all leave the same kind of
label, the results stated in an introduction, so their deviations cannot come from different
purposes, and they are stable. The subjective part of interestingness is therefore partly a
matter of taste, although I cannot say how much of the differences between signallers is.

\section{Other corpora}
\label{app:corpora}

This appendix describes the corpora of \cref{sec:libraries}, the arXiv papers and the libraries of Metamath and HOL Light, and provides the details of the results obtained on them.

\subsection{arXiv}
\label{app:arxiv}

From the LaTeX sources of arXiv papers
\citep{redpajama2023} I extract the theorem, proposition, corollary and lemma environments, with
the section they appear in and, where it can be found, their proof. I find the proof either directly after the result or through a phrase such as ``Proof of
Theorem~\textbackslash ref\{\ldots\}''. I sample 20{,}144 papers from 2010--2018, with up to 750
per category among the 20 largest mathematics categories and up to 1{,}250 per category at the border of mathematics
(mathematical physics, theoretical computer science and statistics theory), for 384k results including all categories. I keep the features that carry over from \mathlib{}. These are the length of the statement,
its hypotheses, the share of it written in mathematical notation, its surprisal to a language model (\emph{Qwen3-8B-Base} and
\emph{Qwen2.5-Math-7B} \citep{yang2024qwen25math}, alone and in context), and its novelty with respect to earlier papers. For
results with a proof, they also include the length of the proof, the proof/statement ratio, the
citations and internal references in the proof, and the height of the chain of results of the same
paper that the proof rests on. Dependencies between papers are not available, and neither are the
features built on them, such as the entropy of the subfields a proof draws on and usage.

\paragraph{Sources.} The metadata come from a CC0 snapshot of arXiv metadata (1.28M papers with a mathematics or border category), and the \LaTeX{} sources from the arXiv split of RedPajama \citep{redpajama2023}, with preambles and bibliographies stripped. Citation counts come from Semantic Scholar, which finds 99.8\% of papers by their arXiv identifier and agrees with OpenAlex counts on journal DOIs (Spearman correlation 0.945).

\paragraph{Extraction.} I extract the environments whose name matches theorem, proposition, corollary, lemma,
conjecture or claim (with common abbreviations and starred forms), with their optional name,
label, section number and statement. A proof is the next \texttt{proof} environment if it
follows immediately, otherwise a \texttt{proof} environment whose optional argument references the
result's label. Results whose name contains a citation are restatements of known results and are
excluded. In the analysis, a later result whose token set overlaps a main result's by more than
80\% (Jaccard) is treated as its restatement and dropped (1.8\% of candidates).

\paragraph{Sample.} A paper is eligible if it dates from 2010--2018 and has at least three sections, at least three results, at least one main result and at least one other result, which leaves 68{,}889 papers. I draw up to 750 papers per category among the 20 largest mathematics categories, and up to 1{,}250 per border category (math-ph, cond-mat.stat-mech, cs.IT, cs.CC, cs.DM, cs.DS, cs.LO, math.ST). This yields 20{,}144 papers and 418k results, of which I score the 384k statements of 20 to 4{,}000 characters. The category cond-mat.stat-mech contributes only 11 eligible papers.

\paragraph{Features.} Surprisal uses the header ``\texttt{\% A result from a mathematics research
paper.}'' followed by \texttt{\textbackslash begin\{theorem\}} (bare), or the preceding text of the
paper (in context). The judge prompt is the \mathlib{} prompt adapted to \LaTeX, with citations
replaced by \texttt{[ref]} and the environment's name never shown. Height is the longest chain of
internal references among results with a proof.

\paragraph{Evaluation.} I compute the AUROC within each paper, average it over
papers, and obtain confidence intervals by resampling papers.

\paragraph{A harder comparison.} Without formal proofs, the features are weaker than the judge on \Lintro{} (\cref{sec:arxiv}). The comparison is also harder, since a main theorem is
compared with the other theorems and propositions of its paper, which are themselves selected
results, whereas a main result of \mathlib{} is compared with every lemma of its subfield.

\paragraph{Recovering dependencies.} The dependencies of an informal proof are mostly implicit, so the features that need them are missing. Recovering the dependencies between informal results, as my links from later citations to numbered results (\cref{app:forward}) begin to do, would close part of the gap between the features and the judge.

\paragraph{Border categories.} As in \mathlib{} (\cref{app:core}), I also check separately the categories at the border of mathematics, which are the most likely to value different aspects. Their own weights do not do better either. The differences (mathematical physics $-0.014$, theoretical
computer science $+0.004$, statistics theory $+0.001$) are of the size found for groups of four other categories drawn at random, similarly to the central groups of \mathlib{}.

\subsection{Metamath and HOL Light}
\label{app:libraries}

I chose these two libraries because the results a proof uses can be read from its text without running the prover, and because their logics and communities differ from those of \mathlib{}. Metamath works in a minimal logic, and HOL Light is a long-standing system in higher-order logic, used for the formal proof of the Kepler conjecture \citep{hales2017formal}. In both, a proof lists or mentions the results it uses, so I can compute
statement size, hypotheses, proof size, the proof/statement ratio, dependency height, the entropy
of the areas a proof draws on, and the number of users. The set.mm library of Metamath contains 28{,}229 theorems with explicit proofs in a minimal logic. The library of
HOL Light \citep{harrison2009hollight} contains 34{,}259 theorems, whose proofs are scripts of
tactics in OCaml.

\paragraph{set.mm.} I use Metamath's set.mm at commit 7d022502 of its develop branch, without users' mathboxes and discouraged duplicates, which leaves 28{,}229 theorems in 18 parts and 622 sections. Each compressed proof lists the statements it uses, so no build is needed. There are four labels. The first is Metamath's ``100 theorems'' list (74 positives). The second, \emph{textbook}, marks the theorems whose comment cites a textbook statement or page (2{,}285). The third, \emph{other names}, marks those whose comment names the result, but not after a person (1{,}317), and the fourth, \emph{eponym}, those whose comment names the result after a person (138). A capitalized word counts as a name if set.mm's comments almost never write it in lower case, without counting words that begin a sentence, and the authors of cited textbooks (``Enderton's theorem'') are excluded. Without these two rules, 65 further comments, such as ``Contrapositive inference for inequality'', would count as eponyms. Between signallers, matched minus mismatched AUROC ranges from $+0.042$ to $+0.383$ on the 100 list, from $+0.048$ to $+0.140$ on textbook results, from $-0.010$ to $+0.311$ on eponyms and from $+0.247$ to $+0.333$ on other names.

\paragraph{HOL Light.} I use HOL Light at commit d30e823, which contains 46{,}624 theorems proved with \texttt{prove}, in 31 directories. The dependencies of a theorem are the names of theorems and definitions that its tactic text mentions, resolved first in the same file, then in the rest of the library. I call \emph{library proper} the library without the files written to prove items of the list, examples, tutorials and stand-alone tools, which contains 34{,}259 theorems. The first label marks the theorems named on the HOL Light ``100 theorems'' page (81, of which 27 are in the library proper). The other two, other names and eponyms, come from the comment that precedes the theorem (90 and 55 positives). Between signallers, matched minus mismatched AUROC ranges from $-0.051$ to $+0.026$ on the 100 list, from $-0.007$ to $+0.068$ on eponyms and from $+0.096$ to $+0.151$ on other names. With all theorems, the ex-ante AUROC on the 100 list is 0.736. Hypotheses (implications in a HOL term) and the entropy of areas (over 31 directories) are measured differently from \mathlib{}, and their weights change sign, while the weights of the ratio and of height keep theirs.

\paragraph{Transfer of the shared weights.} \Cref{tab:libraries} compares the shared weights of \mathlib{}, applied without refitting, with each library's own models.

\begin{table}[t]
\centering
\caption{Transfer to other libraries. For each label, the pooled AUROC (computed over all theorems of the library, without pairing them within subfields) of three models. The first applies the shared weights $\weights_0$ of \mathlib{}'s joint model, fitted in \mathlib{} on the five features the libraries
share, without refitting them on the library. The other two are the library's own models, fitted on the library's label and evaluated on held-out modules, on all its ex-ante features and on the five shared features. In bold, where \mathlib{}'s shared
weights do as well as or better than the library's own model on the shared features.}
\label{tab:libraries}
\small
\setlength{\tabcolsep}{3pt}
\renewcommand{\arraystretch}{1.15}
\arrayrulecolor{tabframe}
\begin{tabular}{|>{\columncolor{tablabel}}l|>{\columncolor{tablabel}}l|c|cc|}
\hline
\rowcolor{tabhead} & & \mathlib{} shared & \multicolumn{2}{c|}{Own model} \\
\hhline{|>{\arrayrulecolor{tabhead}}->{\arrayrulecolor{tabframe}}|>{\arrayrulecolor{tabhead}}->{\arrayrulecolor{tabframe}}|>{\arrayrulecolor{tabhead}}->{\arrayrulecolor{tabframe}}|--|}
\rowcolor{tabhead} & \multirow{-2}{*}{Label (positives)} & weights & All & Shared \\
\hline
set.mm & 100 list (74) & \textbf{0.769} & 0.760 & 0.778 \\
 & other names (1{,}317) & 0.416 & 0.691 & 0.583 \\
 & eponym (138) & \textbf{0.672} & 0.661 & 0.654 \\
 & textbook (2{,}285) & 0.547 & 0.610 & 0.580 \\
\hline
HOL L. & 100 list (27) & 0.686 & 0.711 & 0.739 \\
 & other names (90) & 0.536 & 0.759 & 0.741 \\
 & eponym (55) & \textbf{0.749} & 0.723 & 0.715 \\
\hline
\end{tabular}
\arrayrulecolor{black}
\end{table}

\paragraph{HOL Light's list.} The shared weights of \mathlib{} transfer only partly to HOL Light's ``100 theorems'' list (0.686 against
0.739), which has 27 positives.

\paragraph{Results named after a person.} In set.mm, as in \mathlib{} and HOL Light, most results named after a person are theorems such as Lagrange's and Sylow's theorems, the Cayley--Hamilton theorem or the Bolzano--Weierstrass theorem, and only a minority are laws of logic, such as De Morgan's laws.

\paragraph{Results with other names.} In set.mm, most of them are elementary laws, such as ``Commutative law'' or ``Closure law'', and the shared weights of \mathlib{} tend to rank them below the other results (AUROC 0.416, below the $\tfrac12$ of a random ranking). In HOL Light, they rank them no better than at random (0.536).

\paragraph{Weights of the shared features.} \Cref{tab:libweights} reports the weights of the five shared features, fitted on standardized features: the shared weights $\weights_0$ of \mathlib{}'s joint model, and the weights fitted on the curated labels of set.mm and HOL Light. In every row, the weights of the proof/statement ratio and of dependency height are positive.

\begin{table}[h]
\centering
\caption{Weights of the five features shared by the three libraries, on standardized features. First row: the shared weights $\weights_0$ of \mathlib{}'s joint model, fitted on these five features. Other rows: the weights fitted on a curated label of set.mm or HOL Light.}
\label{tab:libweights}
\small
\setlength{\tabcolsep}{3pt}
\renewcommand{\arraystretch}{1.15}
\arrayrulecolor{tabframe}
\begin{tabular}{|>{\columncolor{tablabel}}l|ccccc|}
\hline
\rowcolor{tabhead} & Statement & Hypo- & Ratio & Height & Area \\
\rowcolor{tabhead} & size & theses & & & entropy \\
\hline
\mathlib{}, $\weights_0$ & $+0.09$ & $+0.30$ & $+0.34$ & $+0.46$ & $+0.16$ \\
set.mm, 100 list & $+0.07$ & $+0.23$ & $+0.17$ & $+1.14$ & $+0.02$ \\
set.mm, eponym & $+0.32$ & $+0.16$ & $+0.37$ & $+0.38$ & $-0.01$ \\
HOL L., 100 list & $+0.88$ & $-0.49$ & $+0.97$ & $+0.35$ & $-0.24$ \\
HOL L., eponym & $+0.87$ & $+0.07$ & $+0.76$ & $+0.09$ & $-0.30$ \\
\hline
\end{tabular}
\arrayrulecolor{black}
\end{table}

\paragraph{Signallers.} The signallers differ accordingly. In set.mm, matched minus mismatched label ranges from $-0.010$ to $+0.383$. The largest gaps involve results with other names, while the ``100 theorems'' list and the eponyms are almost interchangeable ($+0.042$ and $-0.010$, with intervals including 0). In HOL Light, only the results with other names stand apart ($+0.096$ and $+0.151$).

\section{Forward tests}
\label{app:forward}

In \cref{sec:prediction,sec:core,sec:libraries}, every label records selections that were already made when I collected the data, often long before, and many positives have been well known for decades. A profile, and even more a judge that has read about these results, may then recognize how well known a result is instead of how interesting it is. An exploring agent faces a different situation, since it must itself single out novel results, before anyone else has. This appendix provides the details of the three forward tests of \cref{sec:forward}, which predict selections made after the information my models learned from: additions to \mathlib{}'s curated lists, citations of numbered results, and papers posted after the release of the language models.

\subsection{Additions to curated lists}

\paragraph{\mathlib{} history.} From 34{,}769 commits I record the first date each declaration
appears in one of the four lists (each list file parsed at every commit that changed it), and
which current theorems existed at each cut (same file and declared name at the last commit before
the cut; renamed or moved theorems are dropped, about half of each date's theorems are matched).
Features are those of v4.34.1.

\paragraph{Results.} \Cref{tab:x4} reports the results. The profile and the judge learn from the past in different ways. The profile is fitted on past labels, whereas the judge is never fitted on my labels but was trained on texts written before its release. The judge does better than the profile on later additions, and combining the two leads to a performance similar to that of the judge alone. This is expected, since the later additions are mostly classical
results, part of the canon long before \mathlib{} formalized them, and the canon is precisely what the judge knows
(\cref{sec:prediction}). Its top-ranked later additions include Roth's theorem, the corners theorem, the
Erd\H{o}s--Ko--Rado theorem and the Cauchy--Davenport theorem. This test therefore forecasts curation more than the course of mathematical research.

\begin{table}[t]
\centering
\caption{Anticipating additions to \mathlib{}'s curated lists. Models are trained on the lists
as they stood at each cut date and scored on the theorems that existed then and were not listed. A
theorem is positive if it was added to a list later (up to v4.34.1). Pooled AUROC (over all theorems, without pairing them within subfields), with 95\%
intervals from a module bootstrap.}
\label{tab:x4}
\small
\renewcommand{\arraystretch}{1.15}
\arrayrulecolor{tabframe}
\begin{tabular}{|>{\columncolor{tablabel}}l|c|ccc|}
\hline
\rowcolor{tabhead} Cut & Added later & Profile & Judge & Both \\
\hline
2024-01 & 90 & 0.831 {\scriptsize[0.79,0.86]} & 0.889 & 0.883 \\
2024-07 & 104 & 0.853 {\scriptsize[0.82,0.88]} & 0.891 & 0.897 \\
2025-01 & 133 & 0.844 {\scriptsize[0.82,0.87]} & 0.876 & 0.883 \\
2025-07 & 46 & 0.802 {\scriptsize[0.70,0.86]} & 0.845 & 0.845 \\
\hline
\end{tabular}
\arrayrulecolor{black}
\end{table}

\paragraph{Hindsight in the features.} The features, however, are computed on the current version of the library, so a theorem whose statement or proof was edited after the cut date may carry some hindsight. To rule this out, I repeat the test on the theorems whose source text is the same at the cut date and in the current version, from 52\% to 74\% of them depending on the cut. Their current features are those they had at the cut. On them, the profile does slightly better (AUROC 0.82 to 0.88), so hindsight does not explain its success. A theorem is unchanged since a cut if its source text (its declaration line and the indented lines that follow, with whitespace normalized) is the same at the last commit before the cut and in v4.34.1. Restricted to unchanged theorems, for training and test, the profile reaches an AUROC of 0.852, 0.876, 0.880 and 0.817 at the four cuts (23, 32, 54 and 29 later additions), against 0.802 to 0.853 on all theorems, and the judge 0.915, 0.911, 0.905 and 0.877. The judge's pre-training may include \mathlib{}'s list files, which
would favour it.

\subsection{Citations of numbered results}

\paragraph{Citation contexts.} To observe citations of single results, I use the arXiv papers of \cref{sec:arxiv}. For the 18k of them that Semantic Scholar identifies by their arXiv identifier, I take the sentences of later papers that cite them (487k such citation contexts), and keep those in which the only work cited is the paper in question, together with the number of one of its results, as in ``[12, Theorem 3.2]'' or ``Theorem 3.2 of [12]'', so that the number can only refer to a result of that paper.

\paragraph{Numbering schemes.} To
know which result of the cited paper ``Theorem 3.2'' is, I need its numbering scheme, for
instance whether results are numbered within sections or throughout the paper, and the source I
have lacks the preamble that sets it. I therefore choose the scheme under which the citations land
on results of the cited kind (a ``Theorem 3.2'' on a theorem). On held-out citing papers, this
choice puts the cited number on a result of the cited kind 64\% of the time, against 35\% if it pointed to a random result of the paper (details below).

\paragraph{Numbered citations.} Semantic Scholar provides citation contexts for all 18{,}361 papers with
an arXiv identifier (487{,}233 contexts). A context counts if every citation marker in it names the
cited work and it states a result kind and number (``[12, Theorem 3.2]'', ``Theorem 3.2 of
[12]''): 16{,}032 contexts, 10{,}660 citing and 5{,}156 cited papers. RedPajama strips preambles,
so each result's printed number is computed under eight schemes (per section or global; one shared
counter, one for theorems only, or one per kind; with or without equations; per subsection) plus a
separate counter for lettered environments and appendices; 99.6\% of results receive a number.
Each paper's scheme is the one under which its references land on results of the cited kind.
I call \emph{kind-blind} the schemes in which results of different kinds share a counter, since only for them is landing on the cited kind informative. Held out one citing paper at a time, kind-blind schemes agree on 64\% of 9{,}417 references
(chance 35\%). 10{,}223 of 14{,}727 references (69\%) resolve to a result. Restatements of an
introduction result later in the paper are merged into it (a citation of either counts). The
papers whose numbering is unambiguous, 1{,}128 of them, are those with a kind-blind scheme and at least two references, all resolved.

\paragraph{Cited results.} With these schemes, 8{,}174 results of the sampled papers are cited by number in at least one later paper. Nearly all of them lie in 4{,}244 papers that also have uncited results. These papers contain 96k results in total, of which 8.5\% are cited.

\paragraph{Authors as a baseline.} Being stated in the introduction predicts later citations. This is expected, as many readers read only the introduction of a paper, which is written to present the results meant to be remembered, used and built upon, and they rely on the authors' judgment of which results these are. Being stated in the introduction reaches an AUROC of 0.588 for predicting later citations, the kind of result (theorem, then proposition, corollary and lemma) 0.611, and the two combined 0.625.

\paragraph{What the predictors see.} Over all results, the judge, which sees only the statement, and the features of \cref{sec:arxiv}, which never include the section or the kind of result, also predict citations. Among the results stated in the introduction, which are equally visible, features and judge still predict which ones get cited (0.573).

\paragraph{Citing authors and authors.} Citing authors value the same properties of a result as the authors, with a small deviation of their own. I fit the features and the judge twice, once to predict which results get cited and once to predict which results the authors state in their introduction, and use both to predict citations. The first does better by $+0.022$ $[+0.017, +0.028]$, but the second keeps most of the signal (0.579 against 0.601). Citing authors and authors thus behave like the signallers of \mathlib{}.

\paragraph{Unambiguous numbering.} On the 1{,}128 papers whose numbering is unambiguous, the results are nearly the same. The authors' choices reach 0.621 (against 0.625 on all papers), features and judge 0.602 (against 0.601), their gain over the authors' choices is $+0.016$ (against $+0.025$), and the advantage of the weights fitted on citations is $+0.017$ (against $+0.022$).

\subsection{Papers posted after the language models}

\paragraph{Contamination.} A judge may find the main
results of a paper because it has read the paper beforehand, which is a form of data contamination \citep{golchin2024time}. The same holds for the profile, whose surprisal features come from language models. To test this, as in studies of contamination across training cutoffs \citep{roberts2024cutoff}, I collect 3{,}150 papers posted
from November 2025 to August 2026, after the release of the judge (\emph{Qwen2.5-32B}, 2024) and
of the models used for surprisal, with the same extraction, sampling rules and features. Among them,
1{,}418 papers have both results stated in the introduction and other results.

\paragraph{Post-cutoff papers.} I fetched from arXiv the e-prints of 150 papers per mathematics category of the main sample, posted
from 2025-11 to 2026-08. The main file of a paper is the one with
\texttt{\textbackslash documentclass}, with \texttt{\textbackslash input} and
\texttt{\textbackslash include} expanded and the preamble dropped, as in RedPajama. 3{,}150 were
fetched, 1{,}504 meet the sample rules, and 1{,}418 of them have both results stated in the introduction and other results (21{,}532 candidate results). Novelty, which needs earlier papers of
the same period, is omitted on both sides of the comparison.

\paragraph{Method.} On each period separately, I score the judge and fit the profile with cross-validation by paper, as in \cref{sec:arxiv}. If either succeeded by remembering papers, it would do worse on the recent period.

\paragraph{Results.} For the label \Lintro{}, the judge reaches 0.702 $[0.687, 0.714]$ against 0.674, a difference of $+0.027$ $[+0.013, +0.039]$. The profile reaches 0.590 against 0.526, a difference of $+0.064$ $[+0.049, +0.080]$. Weights fitted on the older papers and applied unchanged to the recent ones reach 0.587, against 0.590 for weights fitted on the recent papers, a difference of $-0.003$ $[-0.011, +0.004]$.

\paragraph{Why both do better on recent papers.} The two samples also differ in period and composition. The older papers come from RedPajama, which leaves out the files a paper includes from others, so their proofs and the references between their results are often incomplete, and the features measured on them are noisier. Weights fitted on them still capture the profile, and they rank the recent papers better because these are complete e-prints, whose features are measured more accurately.

\section{Beyond the current features}
\label{app:limits}

This appendix provides the details of \cref{sec:limits}: the settings I consider, the tests on definitions, the history of a result, the distance between the subfields a result connects, the construction and analysis of the close variants of a theorem, and how an explorer could use interestingness given these limits.

\subsection{Settings}

I look for aspects of interestingness that the features cannot capture in harder settings. Definitions, first, may be valued for what they make possible more than
for their own form, and what they make possible lies outside the ex-ante features. An agent must also choose between a theorem and close variants of it, such as slightly more or less general versions. These variants are distinct statements, but they share most of their structure. An agent that explores produces many of them. A good predictor should pick the actual theorem, the one mathematicians kept. Since every variant in my test is proved, this choice does not depend on what can be proved, only on what mathematicians value in a statement. Finally, I consider the distance between the subfields a result connects through its statement and its proof. Results that connect subfields far apart, which few mathematicians master together, are often highly valued (\cref{app:framework}), and the profiles I considered so far do not account for this distance.

\subsection{Definitions}

In Lean, a definition has a type, which says what kind of object it defines (for instance a property of natural numbers), and a value, the expression that defines it. I compute on each definition the same features as for theorems, with the type in the role of the statement and the value in the role of the proof.

\paragraph{Transfer from theorems.} I fit the ex-ante and usage features together on the theorems of \Lmain{}, and apply these weights unchanged to definitions. They reach 0.522, barely above chance, against 0.682 for the same features fitted on definitions. The two models weigh the features differently. Usage, which gets almost no weight for theorems ($+0.02$), dominates for definitions ($+0.69$), and the weights of novelty and of
the entropy of the subfields a definition draws on change sign. The shared weights of \cref{sec:core} are thus weights for results, and judging a new concept
requires seeing it used.

\paragraph{Anticipating use.} I also test whether the ex-ante features anticipate how much a definition will be used. From \mathlib{}'s git history, I count how many times the name of each definition appears in the library on 1 July 2024 and in v4.34.1. Among the 7{,}577 definitions that existed on 1 July 2024, I predict which ones are among the 10\% most used in v4.34.1, and which ones are among the 10\% whose use grew most since. Once their use in July 2024 is known, the ex-ante features add nothing to the prediction of their later use ($-0.002$), and at most $+0.013$ $[-0.004, +0.028]$ to the prediction of their growth. The form of a definition does not anticipate its use.

\subsection{History}

A theorem that settles a problem left
open for a long time, such as Fermat's Last Theorem, is valued in part for that history, and the
longer the problem stayed open, the more the result is valued. This information is available as
soon as the result is proved, since an agent that settles a known conjecture knows it, but it lies
in the history of the community: which problems were posed, when, and how many attempts failed before the result. None of the features I defined records this history, since the features describe the statement, the proof and the place of the result in the library as it stands today. It is part of what the
judge knows about the canon, and it may explain part of the judge's lead on curated lists, which
contain many results of this kind. A feature recording the history of a result, such as whether it settles a known problem and how long that problem had been open, could be added to the profile. Lists of open problems, such as the database of Erd\H{o}s problems, would tell whether a result settles a known problem, and the date at which the problem was first posed would tell how long it stayed open.

\subsection{Distance between subfields}

\paragraph{Two measures of distance.} A result that bridges subfields so far apart that few people master both is often highly valued \citep{gangloff2026creativity}, and \citet{patel2026interesting} note that the proof/statement length ratio does not capture this. Intuitively, the farther apart two subfields are, the rarer and the more valued a connection between them. The shared
weights reward a proof that draws on \emph{several} subfields, as measured by the entropy of the subfields
of the results it uses, but this entropy is the same whether the subfields are close or far apart. I
therefore add to the profile the \emph{distance} between the subfields a result draws on through its statement and its proof, measured in two independent
ways. The first is how rarely \mathlib{} uses the two subfields together, and the second is how rarely
arXiv papers are cross-listed in both. The first measure is computed both for the 32 subfields and for the 1{,}404 directories one level below them (such as Analysis/SpecialFunctions). The two measures agree little, with a Spearman correlation of 0.20.

\paragraph{Evaluators reward results that draw on several subfields, whether close or far apart.} Once the profile
is known, distance adds nothing for any signaller. The differences range from $-0.006$ to $+0.004$, and
none is above zero beyond its interval. This agrees with what has been found for citations in mathematics \citep{luo2026holes}. Taken alone, distance ranks the results singled out by signallers only slightly above the others (AUROC 0.56 to 0.65), and part of this comes from its correlation with the number of subfields (Spearman correlation 0.32 to 0.66).

\paragraph{Interpretation.} This does not mean that distance plays no role. Both measures are computed after the fact, and a connection that is used often makes the two subfields it connects look close, so the measures are low precisely where a valued connection was built. Bridging in
the sense of a \emph{first} connection between two subfields needs the date of each connection, and
remains untested.

\subsection{Close variants of a theorem}
\label{app:pipeline}

\paragraph{Procedure.} Each variant is elaborated in Lean and proved by an open prover, and trivial reformulations are excluded. Each result and its proved variants form a group, with a median of 6 candidates. I fit weights on 5{,}502 pairs of a real result and a variant, built from other results, and use them to rank the candidates of each group. The steps are as follows.

\begin{enumerate}
  \item \textbf{Pool.} 1{,}500 labelled theorems whose statement round-trips through the
    pretty-printer, each with a matched real lemma from the same module (87\% of the time).
  \item \textbf{Variants.} Qwen2.5-32B-Instruct writes 8 variants per theorem, 2 of each type:
    specialization, weakening, trivial variant, generalization (21{,}486 in all).
  \item \textbf{Elaboration.} Lean elaborates each variant in the original's context; 13{,}420
    compile and are not definitionally equal to the original.
  \item \textbf{Truth.} DeepSeek-Prover-V2-7B, $k{=}8$, with the original as a hint: 6{,}238
    proved (generalization 0.25, specialization 0.51, trivial variant 0.58, weakening 0.48).
  \item \textbf{Difficulty.} The fraction of 32 attempts that succeed (pass@32) for proved variants without the hint and with the original
    forbidden, matching the setup for real results.
  \item \textbf{Groups.} All items in the same format, scored at the original's position: 1{,}467
    groups, median 6 candidates. The top-1 accuracy of a ranker is the fraction of groups in which it ranks the real result first, and its mean reciprocal rank (MRR) is the mean of the inverse of the rank of the real result. A random choice has a top-1 accuracy of 0.21 and an MRR of 0.45.
\end{enumerate}
The weights are fitted on a separate \emph{contrast set}, built with the same pipeline on 1{,}993 other labelled theorems: 28{,}569 variants,
17{,}808 elaborated, 8{,}335 proved, leaving 5{,}502 non-trivial real-vs-variant pairs. The model
is a conditional logit model, which compares each real result only with the variants of its own group.

The weights fitted in \cref{sec:prediction} do poorly on whole groups made of a real result, a real lemma and variants. The best of them reach a top-1 accuracy of 0.28 to 0.29, below the judge (0.34) and context surprisal alone
(0.39). These weights are dominated by statement length, which separates main results from small
lemmas across the library but points the wrong way inside a group, where weakenings and
generalizations are longer than the real result. I therefore compare real results separately with lemmas and with variants.

\begin{table}[h]
\centering
\caption{Ranking a real result among its close variants. Left: rankers, and the profile fitted on the contrast set with some features removed (top-1 accuracy and MRR over all
groups, and top-1 accuracy with trivial variants removed, the measure reported in \cref{sec:rediscovery}). Right: probability that the real result beats a variant of the held-out
type, when that type is left out of the fit.}
\small
\begin{tabular}{@{}lccc@{}}
\toprule
Ranker & Top-1 & MRR & Top-1, no trivial \\
\midrule
Random & 0.21 & 0.45 & 0.26 \\
Profile of \cref{sec:prediction} (\Lmain{} weights) & 0.28 & 0.52 & 0.37 \\
Judge & 0.34 & 0.59 & 0.45 \\
Profile fitted on the contrast set & 0.55 & 0.75 & 0.62 \\
\quad context surprisal only & 0.54 & 0.73 & \\
\quad without context & 0.33 & 0.60 & \\
\quad without context and binders & 0.30 & 0.55 & \\
\bottomrule
\end{tabular}
\hspace{1em}
\begin{tabular}{@{}lcc@{}}
\toprule
Held-out type & All features & No context, no binders \\
\midrule
Specialization & 0.57 & 0.16 \\
Weakening & 0.90 & 0.55 \\
Generalization & 0.87 & 0.56 \\
\bottomrule
\end{tabular}
\end{table}

\paragraph{The leak in the file context.} On all groups, the features measuring the surprisal of the statement given the preceding text of the file rank the real result first in 54\% of groups, against 55\% for the full profile. A second comparison, with lemmas, isolates this effect. I pair the same results with real lemmas of the same module, mostly matched in size (\cref{tab:e3a}). When both the real result and the lemma are scored given the text that precedes the real result, the real result is less surprising in 70\% of pairs. When each is scored given the text that precedes it, the real result is less surprising in only 54\% of pairs. The text before a result thus makes that result expected there, whatever its interest. Against lemmas, this leak does not explain the success of the profile. When each statement is scored given its own preceding text, the full profile still ranks the real result above the lemma in 59\% to 62\% of pairs (judge 65\%, random 50\%), so it can tell a result that humans singled out from an ordinary lemma.

\begin{table}[h]
\centering
\caption{Probability that a real result is ranked above a matched real lemma of the same module. Rows: whose
result; columns: whose profile.}
\label{tab:e3a}
\small
\begin{tabular}{@{}lccc@{}}
\toprule
Result labelled by & \Lmain{} prof. & \Llists{} prof. & \Lnamed{} prof. \\
\midrule
\Lmain{} ($n{=}923$) & \textbf{0.583} & 0.551 & 0.532 \\
\Llists{} ($n{=}114$) & 0.586 & \textbf{0.602} & 0.575 \\
\Lnamed{} ($n{=}474$) & 0.527 & 0.588 & \textbf{0.646} \\
\bottomrule
\end{tabular}
\end{table}

\paragraph{Do the weights generalize across types of variants?} Without the context features, the profile ranks the real result first in 33\% of groups, against 21\% for a random choice (\cref{sec:rediscovery}). This remaining gain may reflect a preference for the real result, or only the habits of the language model that writes the variants. If the weights captured what makes the real result preferable, this would hold against any type of variant, including a type they were not fitted on. If they only captured how each type of variant differs from the real result, they would fail on a type they have not seen. I therefore fit them on all types of variants but one, and test them on pairs of a real result and a variant of the left-out type. I also remove the counts of arguments, which directly reveal how a variant was built. The weights pick the real result in 55\% of pairs for weakenings and 56\% for generalizations, close to the 50\% of a random choice, and in only 16\% for specializations, which they prefer to the real result. They thus learn how the language model builds each type of variant, for instance that a specialization has fewer arguments, and not what makes the real result preferable.

\subsection{Using interestingness in an explorer}

My results suggest how an
explorer could use interestingness. For theorems, a profile of ex-ante features can be fitted once, on the selections of one library, and its shared weights used by default (\cref{sec:core,sec:libraries}). When the explorer works for a given community or person, a small
deviation fitted on their selections adds the rest. Combined with a judge, which carries most of the
signal where dependencies are implicit, it can rank the results an explorer produces as soon as they
are proved. The profile and the judge are not enough, however, for two types of decisions that an explorer must make, choosing among close variants of a theorem and judging a new definition. To choose among variants, the explorer would have to compare them with each other, for instance by what each lets it prove. Since a definition is valued for its use (\cref{sec:limits}), an explorer could keep a new definition provisionally, use it to state and prove results, and
judge it by this use.

\section{Further related work}
\label{app:related}

This appendix describes in more detail the work summarized in \cref{sec:related}: how automated discovery has selected what it produces, how exploring agents and systems for automated mathematics implement interestingness, how my work differs from them, and other accounts of interestingness in mathematics and comparisons of human and model judgments.

\paragraph{Selection in automated discovery.} The question of what to keep is as old as
automated discovery. Early systems of automated mathematics, such as Lenat's Automated Mathematician (AM)
\citep{lenat1976am}, Fajtlowicz's conjecture generator Graffiti \citep{fajtlowicz1988graffiti} and
Colton's theory-formation system HR \citep{colton2000interestingness}, scored new concepts and
conjectures by hand-designed measures of interest.
Reinforcement learning rewards curiosity, learning progress or novelty
\citep{schmidhuber2010creativity,oudeyer2007intrinsic,pathak2017curiosity,lehman2011novelty}.
Open-ended learning now queries a foundation model to decide which problems a person would find interesting
\citep{zhang2024omni,faldor2025omniepic}, and open-endedness has been argued to be essential for
systems that go beyond human knowledge \citep{hughes2024openended}. Agents for research and for
mathematics already generate ideas, conjectures and lemmas faster than anyone can read them
\citep{lu2024aiscientist,poesia2024minimo,leanconjecturer2025,long2026meca,chen2026moonshine}. Current systems rely on a single proxy, such as the ratio of proof length to statement length
\citep{patel2026interesting}, being barely provable for the current prover
\citep{poesia2024minimo,dong2025stp}, compression
\citep{bengio2024aimathematician,herrmann2026interestingness}, or the verdict of a language model
\citep{zhang2024omni,long2026meca}. The paragraphs below describe these systems in more detail.

\paragraph{Exploring agents.} In reinforcement learning, intrinsic motivation, a notion borrowed from psychology, provides an agent with a reward of its own, in addition to the reward associated with the problem. This reward defines what the agent finds interesting, for instance what makes progress in compressing or learning from its observations, or what it fails to predict
\citep{schmidhuber2010creativity,oudeyer2007intrinsic,pathak2017curiosity}. \citet{schmidhuber2010creativity} calls an observation interesting when it lets the agent compress its history better. Novelty search, in evolutionary algorithms, drops the reward associated with the problem altogether, and keeps the candidate solutions whose behaviour differs most from the behaviours found so far \citep{lehman2011novelty}, so that what is interesting is what has not been seen before. In both cases, interestingness is defined relative to the agent and its history, and says nothing of what people would find worth pursuing. Other agents instead query a foundation model, used as a model of human notions of interestingness, whether to choose the problems they learn to solve next, as in OMNI and OMNI-EPIC \citep{zhang2024omni,faldor2025omniepic}, or to judge their own research ideas, with reviewers based on large language models (LLMs) \citep{lu2024aiscientist}. Experts rate the ideas of LLMs as more
novel but slightly less feasible than their own \citep{si2025ideas}.

\paragraph{Systems for automated mathematics.} AM, the Automated Mathematician \citep{lenat1976am}, explored elementary set theory and number theory with hand-written heuristics that rated how interesting each concept was, for instance from the number of its examples or from its links to concepts already rated interesting, and chose what to explore next from these ratings. It rediscovered prime numbers in this way. Later, EURISKO represented its heuristics as concepts of the same kind as those it explored, so that heuristics could also create, modify and rate other heuristics, for instance by specializing a heuristic that had often led to interesting concepts \citep{lenat1984am}. Graffiti kept the conjectures that were
not implied by earlier ones \citep{fajtlowicz1988graffiti}, and HR scored concepts and conjectures
by novelty, surprisingness, applicability and comprehensibility
\citep{colton2000interestingness,colton2002book}.

Recent systems mostly rely on single criteria, such as the length of a proof relative to its statement \citep{patel2026interesting}, or being barely provable for the current prover \citep{poesia2024minimo,dong2025stp}. The proof/statement ratio is a single criterion, which \citet{patel2026interesting} validate against the compression of the library that a theorem enables. LeanConjecturer generates conjectures in Lean and keeps those that are new and that a prover can prove \citep{leanconjecturer2025}. Some other research agents aim at conjectures that are ``valuable'' or ``of
mathematical significance'', with critic agents that assess research value
\citep{long2026meca,chen2026moonshine}. COMPOSE conditions a language model on two graphs, the citations between arXiv papers and the dependencies between the theorems of \mathlib{}, to generate theorems that future papers may state, and evaluates them by how well they match papers posted later \citep{busbib2026compose}. Interestingness is then judged by plausibility, that is, by whether a result fits the direction that research is taking.

Instead of fixing a criterion by hand, \citet{tsoukalas2025learning} search for a measure of interestingness. In an environment where an agent builds a theory from basic definitions, a language model writes and evolves programs that rate how interesting a concept is, and keeps those that lead the agent to discover more concepts. Although the measure is learned, interestingness is still judged by a single criterion, the production of concepts.

\paragraph{How my work differs.} Most systems for automated mathematics rely on a single criterion, whereas I combine several criteria, including some of theirs, as features of one profile. Like \citet{tsoukalas2025learning}, I learn how much each aspect matters, but I learn it from the selections that mathematicians make in their practice, whereas they learn it from what their system produces. My features are chosen by hand, which is less systematic than a search over programs, but each of them has a clear meaning, so the fitted weights indicate which of my intuitions about interestingness these selections support.

\paragraph{Accounts of interestingness in mathematics.} \citet{bengio2024aimathematician}
characterize valuable mathematics by short descriptions close to many provable statements.
\citet{herrmann2026interestingness} view the interestingness of an object as a prediction of how much studying it will help an agent compress what it observes, in the line of \citet{schmidhuber2010creativity}, and study when such a prediction is possible. \citet{aksenov2026compression} characterize human mathematics, within formal mathematics, by its high compressibility through nested definitions and lemmas. They represent results as strings of primitive symbols, in which definitions and theorems act as macros, that is, names that stand for longer strings. A useful macro shortens many strings at once, and macros can be nested. In \mathlib{}, they observe that the written length of a result stays roughly constant, while the length of its full expansion into primitive symbols grows exponentially with the depth of nesting. \citet{barkeshli2026structure} share this view, and turn it into criteria for an agent that makes discoveries. One of these criteria requires the agent to justify which results it selects, for instance by how much a result shortens the proofs of others. Where Aksenov et al.\ measure how much earlier definitions and lemmas shorten a result, this criterion measures how much a result shortens the others. Network analyses of \mathlib{}, however, find that usage counts measure technical utility more than depth \citep{mathlibnetwork2026}. This agrees with my finding that usage adds no information for theorems, while I find the reverse for definitions. Graphs of results are not limited to formal libraries. TheoremGraph \citep{kurgan2026theoremgraph} and Connected Theorems \citep{berczi2025connected} build them from papers. Like the systems for automated mathematics described above, Aksenov et al.\ measure interest by a single quantity, but they do not use it as a mere proxy. They derive this quantity from what characterizes human mathematics, which is what interestingness should help an agent imitate. I also try to characterize interestingness precisely, but I do not expect a single quantity with a clean mathematical definition to capture it. I therefore learn it from the selections of mathematicians, as a combination of features that includes close counterparts of their quantity, dependency height and unfolded size.

\paragraph{Human and model judgments.} Other studies compare the judgments of humans and of LLMs on what is interesting. For instance, \citet{matterofinterest2025} find that LLMs do not robustly match human judgments of the
interestingness of mathematical problems, and \citet{gu2024scimuse} find that GPT-4o predicts which research ideas experts rate as interesting about as well as a model trained on these ratings (AUROC about 0.67 against 0.64), although the trained model is more precise at the top of its ranking. Reward models learn what people prefer from their choices between outputs \citep{christiano2017deep,ouyang2022training}, as I learn what mathematicians value from what they single out. Closest to my
approach, \citet{quaresma2026geometric} survey experts on what makes geometric theorems interesting, and report six dimensions, namely simplicity and clarity, surprise, conceptual depth, usefulness, aesthetic value, and historical or pedagogical relevance. I also study the judgments of experts, but through the traces they leave in research artefacts, and I learn which combination of measurable aspects best accounts for what they select.

\end{document}